\documentclass[pdflatex,sn-nature]{sn-jnl}

\usepackage{graphicx}%
\usepackage{multirow}%
\usepackage{amsmath,amssymb,amsfonts}%
\usepackage{amsthm}%
\usepackage{mathrsfs}%
\usepackage[title]{appendix}%
\usepackage{xcolor}%
\usepackage{textcomp}%
\usepackage{manyfoot}%
\usepackage{booktabs}%
\usepackage{algorithm}%
\usepackage{algorithmicx}%
\usepackage{algpseudocode}%
\usepackage{listings}%

\theoremstyle{thmstyleone}%
\theoremstyle{thmstyletwo}%

\theoremstyle{thmstylethree}%

\begin{document}

\title{WildLIFT: Lifting monocular drone video to 3D for species-agnostic wildlife monitoring}


\author[1,2]{\fnm{Vandita} \sur{Shukla}}\email{vshukla@fbk.eu}

\author[1]{\fnm{Fabio} \sur{Remondino}}\email{remondino@fbk.eu}

\author[3,4,5]{\fnm{Blair} \sur{Costelloe}}\email{blair.costelloe@ab.mpg.de}

\author*[2]{\fnm{Benjamin} \sur{Risse}}\email{b.risse@uni-muenster.de}

\affil[1] {\orgdiv{3D Optical Metrology}, \orgname{Fondazione Bruno Kessler}, \orgaddress{\city{Trento}, \country{Italy}}}

\affil[2] {\orgdiv{Computer Vision and Machine Learning Systems Group}, \orgname{University of Muenster}, \orgaddress{\state{Muenster}, \country{Germany}}}

\affil[3] {\orgdiv{Department of Collective Behaviour}, \orgname{Max Planck Institute of Animal Behaviour}, \orgaddress{\state{Konstanz}, \country{Germany}}}

\affil[4] {\orgdiv{Centre for the Advanced Study of Collective Behaviour}, \orgname{University of Konstanz}, \orgaddress{\state{Konstanz}, \country{Germany}}}

\affil[5]{\orgdiv{Department of Biology}, \orgname{University of Konstanz}, \orgaddress{\state{Konstanz}, \country{Germany}}}

\abstract{Monocular RGB cameras mounted on drones are widely used for wildlife monitoring, yet most analytical pipelines remain confined to two-dimensional image space, leaving geometric information in video underexploited. 
We present WildLIFT, a computational framework that integrates three-dimensional scene geometry from monocular drone video with open-vocabulary 2D instance segmentation to enable species-agnostic 3D detection and tracking. 
Oriented 3D bounding box labels with semantic face information enable quantitative assessment of viewpoint coverage and inter-animal occlusion, producing structured metadata for downstream ecological analyses. 
We validate the framework on 2,581 manually curated frames comprising over 6,700 3D detections across four large mammal species.
WildLIFT maintains high identity consistency in multi-animal scenes and substantially reduces manual 3D annotation effort through keyframe-based refinement. 
By transforming standard drone footage into structured 3D and viewpoint-aware representations, WildLIFT extends the analytical utility of aerial wildlife datasets for behavioural research and population monitoring.}



\maketitle

\begin{figure*}[ht]
    \centering
    \includegraphics[width=\textwidth]{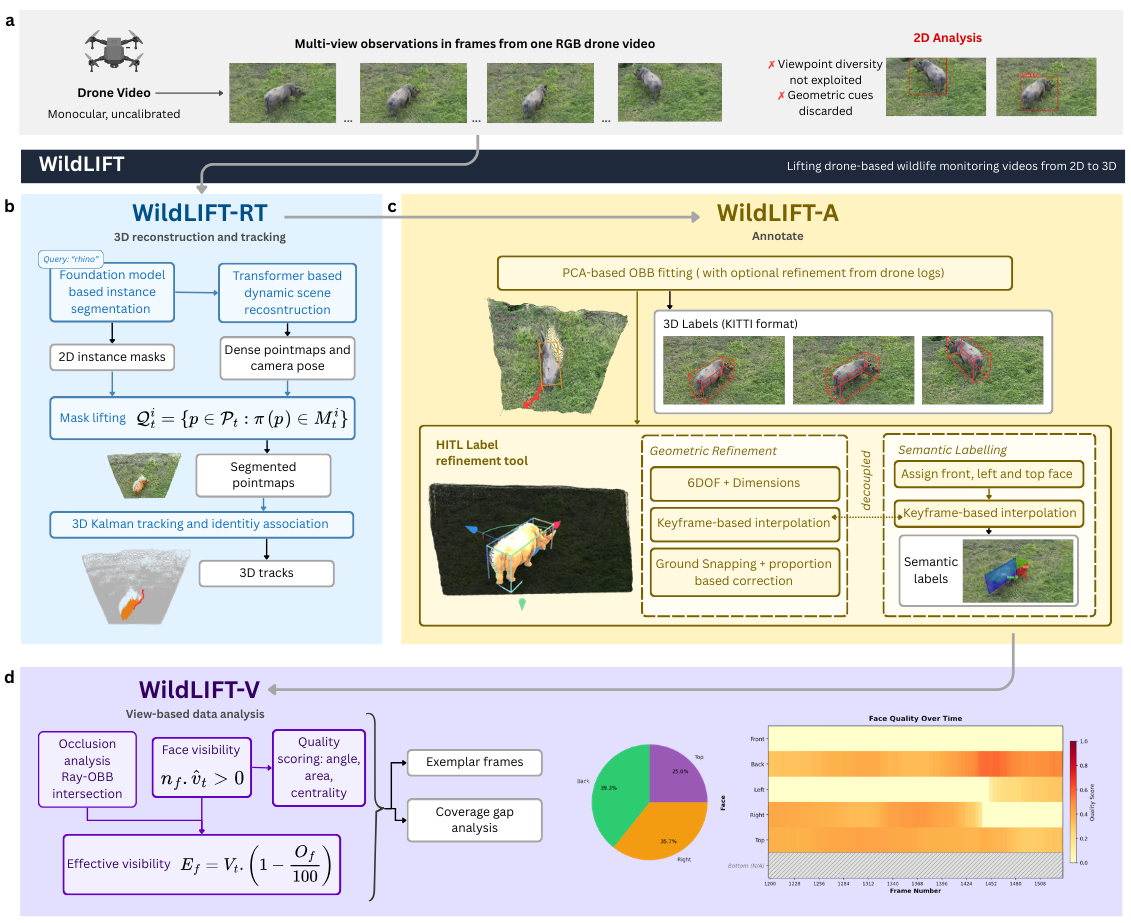}
    \caption{\textbf{The WildLIFT framework lifts monocular drone video from 2D to 3D for wildlife analysis.}
    \textbf{(a)}~Monocular, uncalibrated drone video provides multi-view observations from varying trajectories, yet current analysis discards the geometric cues latent in this data.
    \textbf{(b)}~\textbf{WildLIFT-RT} performs 3D reconstruction and tracking: open-vocabulary instance segmentation produces 2D masks while a feed-forward transformer recovers dense pointmaps and camera poses. Mask lifting ($\mathcal{Q}_t^i = \{\mathbf{p} \in \mathcal{P}_t : \pi(\mathbf{p}) \in M_t^i\}$) projects detections into 3D, and Kalman-filtered tracking associates instances across frames to produce persistent 3D trajectories.
    \textbf{(c)}~\textbf{WildLIFT-A} generates KITTI-format 3D oriented bounding boxes (OBBs) via PCA-based fitting with optional gimbal telemetry constraint. 
    A browser-based human-in-the-loop (HITL) tool supports two decoupled refinement modes: geometric correction (6-DOF adjustment, keyframe-based interpolation via LERP/SLERP, ground snapping) and semantic face labelling (assigning front, top, and left faces; opposing faces are inferred automatically).
    \textbf{(d)}~\textbf{WildLIFT-V} computes viewpoint coverage statistics using the semantic face labels from Stage~2. 
    Per-frame face visibility ($\mathbf{n}_f \cdot \hat{\mathbf{v}}_t > 0$) and quality scoring (viewing angle, projected area, centrality, foreshortening) identify exemplar frames via temporal non-maximum suppression. 
    Ray--OBB intersection tests quantify inter-animal occlusion, yielding an effective visibility score $E_f = V_t \cdot (1 - O_f/100)$ that combines self-visibility with occlusion. 
    Coverage gap analysis flags anatomical aspects never clearly captured. 
    The pie chart summarises the coverage distribution across faces for a single tracklet; the heatmap shows per-frame quality scores over time for each profile.}
    \label{fig:wildlift}
\end{figure*}

Drones have become established tools for wildlife monitoring, enabling surveys of large or inaccessible areas with reduced disturbance to animals~\cite{chabot_wildlife_2015, tuia_perspectives_2022, 
wirsing_rapidly_2022, schad_opportunities_2023, kline_wildwing_2025}.
They complement satellite-based deep learning approaches that operate at landscape scales but lack the temporal resolution and proximity required for individual-level analysis~\cite{wuDeepLearningEnables2023}.
The most commonly deployed sensors are single RGB cameras pre-built into off-the-shelf platforms~\cite{yaney-kellerClosingAirGap2025, 10.1093/biosci/biaf069}.
Although monocular, these cameras combined with drone mobility afford multi-view data acquisition: pilots manoeuvre to follow wildlife and capture data from multiple perspectives~\cite{vogt_drone-based_2025, mesquita_practical_2022}, producing video sequences inherently rich in geometric cues.
Yet current vision-based wildlife analysis, predominantly 2D detection and tracking~\cite{liuDeepLearningMultiple2024, jiang_animal_2022, lauerMultianimalPoseEstimation2022c}, remains confined to image space~\cite{klasen_wildlife_2022}, discarding the three-dimensional scene structure recoverable from the multi-view geometry inherent in drone video.

The recovery of three-dimensional cues from drone video — spatial position, orientation, depth ordering — represents a natural extension of monocular 2D-to-3D lifting, demonstrated for laboratory animals~\cite{gosztolaiLiftPose3DDeepLearningbased2021}, to unconstrained field settings~\cite{xu_animal3d_2023, 10.1093/biosci/biaf069, muramatsu_wildpose_2024, shero_tracking_2021, kulits2025raw, stone_using_2025}. 
Such geometric reasoning would resolve persistent ambiguities in multi-animal movement and mutual occlusion~\cite{hanMultianimal3DSocial2024, klasen_improving_2022}, and is particularly consequential for individual re-identification, where the utility of drone imagery depends not merely on detecting each animal but on capturing diagnostically informative viewpoints of key anatomical regions~\cite{schofield_chimpanzee_2019, schneiderPresentFutureApproaches2019}. 
Photographic identification routinely prescribes specific viewing angles — right-flank images for Grévy's zebras, perpendicular lateral torso views for giraffes~\cite{leeSeeingSpotsQuantifying2018a, bolgerComputerassistedSystemPhotographic2012}, ventral fluke photographs for humpback whales~\cite{cheeseman_advanced_2022}.
This constraint is sufficiently pervasive that viewpoint classification has been formalised as a dedicated filtering stage within detection pipelines~\cite{parhamAnimalDetectionPipeline2018, kuhlAnimalBiometricsQuantifying2013b}. 
For a video sequence, such filtering would operate frame-by-frame discarding images captured from non-prescribed angles even when they depict individuals whose identity could be propagated from diagnostically useful frames elsewhere in the same video sequence.
No existing workflow provides a systematic mechanism to quantify viewpoint distributions across a video, identify coverage gaps, or link an individual's appearance across orientations — capabilities that benefit from integrating 3D scene geometry with instance-level semantic understanding.

Reconstructing consistent 3D geometry from monocular wildlife videos presents significant challenges for traditional 3D computer vision pipelines. 
Monocular videos of dynamic scenes are inherently complex, and unrestricted animal motion within natural environments compounds these difficulties compared to controlled laboratory settings.
Monocular depth networks trained on ground-level data suffer from domain shift when applied to oblique aerial viewpoints~\cite{ranftlRobustMonocularDepth2020, madhuanandSelfsupervisedMonocularDepth2021}, and per-frame inference produces temporally inconsistent geometry under drone motion~\cite{chen2025videodepthanything_cvpr}.
Optimisation-based reconstruction methods that mask dynamic content~\cite{liMegaSaMAccurateFast} fail when animals dominate the frame, as is typical in close-up aerial tracking.
Recent feed-forward reconstruction models that regress 3D geometry directly from video~\cite{wang2025vggt, zhang2024monst3r, cut3r} offer a fundamentally different approach.
Among these, CUT3R~\cite{cut3r} maintains a persistent state representation across frames, producing globally consistent pointmaps without post-hoc alignment while natively handling dynamic scene content.
These properties are well-suited to aerial wildlife footage (Supplementary Fig.~1).

Concurrently, foundation models trained on web-scale data have enabled zero-shot generalisation for vision tasks~\cite{Kirillov_2023_ICCV},including species-agnostic pose estimation across diverse 
taxa~\cite{ye_superanimal_2024}.
Open-vocabulary segmentation methods such as Grounded-SAM~\cite{ren_grounded_2024} provide instance-level detection from natural language queries, enabling deployment to novel taxa by changing a text prompt rather than retraining a species-specific model.
This convergence enables the integration of dense 3D reconstruction from feed-forward models with semantic segmentation from open-vocabulary approaches, providing a framework for 3D wildlife analysis without requiring domain-specific training (see Supplementary Note~1 for extended related work).

Here, we present WildLIFT (Fig.~\ref{fig:wildlift}), a framework that lifts drone videos from 2D to 3D representations through three modules: reconstruction and tracking (WildLIFT-RT), 3D annotation (WildLIFT-A), and viewpoint coverage analysis (WildLIFT-V).
Oriented bounding boxes provide sufficient geometric detail for spatial reasoning and viewpoint characterisation while remaining species-agnostic, unlike parametric shape models that require taxon-specific mesh templates~\cite{zuffi_3d_2017, xu_animal3d_2023}.
As the standard label format for 3D object detection architectures~\cite{geigerAreWeReady2012}, they also position WildLIFT's annotations for use as training data in future supervised 3D detection tasks.

We evaluate WildLIFT on 27 video sequences across four large mammal species (77 individuals, 2,581 annotated frames).
Our 3D tracking achieves perfect recall while attaining the highest identity consistency among tested methods; the annotation tool accepts 93\% of automated OBBs without geometric correction (requiring a mean keyframe ratio of 2.3\%); and the viewpoint module quantifies per-tracklet coverage gaps that are impractical to assess through frame-by-frame inspection.

\section*{Results}

\subsection*{Method overview}
\label{sec:overview}

WildLIFT processes monocular drone video through three sequential modules (Fig.~\ref{fig:wildlift}).
\textbf{WildLIFT-RT} (Reconstruct and Track) integrates feed-forward transformer-based 3D reconstruction with open-vocabulary instance segmentation to produce per-frame pointmaps, camera poses, and identity-consistent 3D tracks.
\textbf{WildLIFT-A} (Annotate) fits oriented 3D bounding boxes (OBBs) to each tracked instance and provides a human-in-the-loop tool for keyframe-based refinement with semantic face labelling.
\textbf{WildLIFT-V} (View) uses the labelled OBBs to compute viewpoint coverage statistics and inter-animal occlusion as structured, queryable metadata.
Each stage consumes the outputs of its predecessor, with only Stage~1 requiring external input (video, a text query specifying the target species, and optionally gimbal telemetry).
The framework also supports independent use: Stage~1 alone suffices for 3D trajectory analysis; Stages~1--2 produce training data for 3D detection models; the full pipeline enables viewpoint-aware data curation for identity-level tasks  (Supplementary Video~1).

\subsection*{Evaluation data}
\label{sec:data}

To assess the framework's generalization across diverse acquisition conditions and taxonomic groups, we compiled an evaluation dataset sourced from archival fieldwork at the Ol Pejeta Conservancy (Kenya), the publicly available KABR dataset~\cite{kholiavchenko_kabr_2024}, and the Bristol Zoological Society. 
In total, this compilation comprises 27 video sequences yielding 2,581 annotated frames of 77 individual animals across four species---rhinos, elephants, zebras, and giraffes---spanning frame rates of 3.75--15~FPS (Supplementary Table~1).
The four target species represent distinct body morphologies necessary for evaluating OBB fitting and tracking capabilities.
Scene complexity ranges from single-animal observations to multi-animal scenarios featuring up to seven individuals.
The successful integration of archival footage demonstrates WildLIFT's robust applicability to legacy data that was not originally collected for 3D analysis.

\subsection*{WildLIFT-RT: 3D tracking in multi-animal scenes}
\label{sec:tracking_results}

CUT3R~\cite{cut3r}, an online feed-forward transformer, recovers dense per-frame pointmaps $\mathcal{P}_t \subset \mathbb{R}^3$ and camera poses $\mathbf{T}_t \in SE(3)$ from uncalibrated video without requiring camera intrinsics (Supplementary Table~2).
Concurrently, Grounded-SAM~\cite{ren_grounded_2024} produces open-vocabulary 2D instance segmentation masks from a text prompt specifying the target species.
Each mask is lifted to 3D by selecting the subset of reconstructed points whose image-plane projections fall within the mask.
A Kalman-filtered tracker with a constant-velocity model associates 3D point clusters across frames using a cost function that combines normalised 3D distance and 2D mask overlap (Equation~\ref{eq:tracking_cost}), solved via the Hungarian algorithm.
A two-tier active/dormant track management scheme handles occlusion: dormant tracks continue to propagate Kalman predictions and can be reactivated upon re-identification.
Full details of the reconstruction backbone selection, tracking formulation, and hyperparameters are provided in the Online Methods (WildLIFT-RT).

\begin{figure*}[ht]
    \centering

    \begin{minipage}[t]{0.02\textwidth}
    \vspace{0pt}\textbf{a}
    \end{minipage}%
    \begin{minipage}[t]{0.97\textwidth}
    \vspace{0pt}
    {\footnotesize
    \renewcommand{\arraystretch}{0.85}
    \begin{tabular*}{\linewidth}{@{\extracolsep{\fill}} ll cccccc}
    \toprule
    \textbf{Species} & \textbf{Method} & \textbf{MOTA}$\uparrow$ & \textbf{IDF1}$\uparrow$ & \textbf{IDS}$\downarrow$ & \textbf{Frag}$\downarrow$ & \textbf{Rec}$\uparrow$ & \textbf{Pred} \\
    \midrule
    \multirow{5}{*}{\shortstack[l]{Giraffe\\(21 GT)}}
     & OC-SORT & 0.951 & 0.866 & \textbf{1} & 13 & 0.952 & 28 \\
     & ByteTrack & 0.981 & 0.920 & 4 & 14 & 0.984 & 27 \\
     & BotSORT & 0.985 & \underline{0.922} & 5 & 14 & 0.989 & 26 \\
     & CUT3R (3D) & \underline{0.990} & 0.918 & 10 & 14 & \textbf{1.000} & 30 \\
     & \textbf{Ours} & \textbf{0.990} & \textbf{0.971} & 9 & \textbf{10} & \textbf{1.000} & \textbf{23} \\
    \midrule
    \multirow{5}{*}{\shortstack[l]{Zebra\\(39 GT)}}
     & OC-SORT & 0.980 & 0.942 & 1 & 12 & 0.980 & 45 \\
     & ByteTrack & 0.994 & 0.955 & 6 & 11 & 0.996 & 43 \\
     & BotSORT & \underline{0.997} & \textbf{0.988} & \textbf{0} & \textbf{3} & 0.997 & 41 \\
     & CUT3R (3D) & \textbf{0.998} & 0.957 & 6 & 9 & \textbf{1.000} & 42 \\ 
     & \textbf{Ours} & \textbf{0.998} & \underline{0.980} & 6 & 6 & \textbf{1.000} & \textbf{40} \\
    \midrule
    \multirow{5}{*}{\shortstack[l]{Elephant\\(8 GT)}}
     & OC-SORT & \textbf{0.998} & \textbf{0.999} & \textbf{0} & \textbf{0} & \textbf{1.000} & \textbf{10} \\
     & ByteTrack & \underline{0.993} & 0.993 & 1 & 1 & \textbf{1.000} & 11 \\
     & BotSORT & 0.934 & 0.891 & 18 & 19 & 0.959 & 30 \\
     & CUT3R (3D) & \underline{0.993} & \underline{0.997} & \textbf{0} & \textbf{0} & \textbf{1.000} & 11 \\
     & \textbf{Ours} & 0.992 & 0.996 & 1 & 1 & \textbf{1.000} & \textbf{10} \\
    \midrule
    \multicolumn{2}{l}{\textit{Rhino (9 GT)}} & \multicolumn{6}{c}{\textit{All methods: MOTA=1.0, IDF1=1.0, Rec=1.0, Pred=9}} \\
    \midrule
    \multirow{5}{*}{\textbf{All (77 GT)}}
     & OC-SORT & 0.977 & 0.936 & \textbf{2} & 25 & 0.978 & 92 \\
     & ByteTrack & \underline{0.991} & 0.956 & 11 & 26 & 0.994 & 90 \\
     & BotSORT & 0.986 & \underline{0.960} & 23 & 36 & 0.990 & 106 \\
     & CUT3R (3D) & \textbf{0.996} & 0.957 & 16 & 23 & \textbf{1.000} & 92 \\
     & \textbf{Ours} & \textbf{0.996} & \textbf{0.982} & 16 & \textbf{17} & \textbf{1.000} & \textbf{82} \\
    \bottomrule
    \end{tabular*}
    }%
    \end{minipage}

    \vspace{8pt}

    \begin{minipage}[t]{0.02\textwidth}
    \vspace{0pt}\textbf{b}
    \end{minipage}%
    \begin{minipage}[t]{0.48\textwidth}
    \vspace{0pt}
    \resizebox{\linewidth}{!}{%
    \footnotesize
    \begin{tabular}{@{}lcccc@{}}
    \toprule
    \multicolumn{5}{@{}l}{\textit{Automated OBB fitting quality}} \\
    \midrule
    Species & Tracklets & Frames & Geo.\ Accept. & Sem.\ Flip \\
    \midrule
    Rhino & 2 & 167 & 167 (100\%) & 10 (6\%) \\
    Elephant & 2 & 181 & 145 (80\%) & 19 (10\%) \\
    Zebra & 4 & 336 & 322 (96\%) & 41 (12\%) \\
    \cmidrule{1-5}
    \textbf{Total} & \textbf{8} & \textbf{684} & \textbf{634 (93\%)} & \textbf{70 (10\%)} \\
    \midrule
    \multicolumn{5}{@{}l}{\textit{Keyframe interpolation efficiency}} \\
    \midrule
    Species & Tracklets & Frames & Keyframes & Time \\
    \midrule
    Rhino & 1 & 84 & 2 (2.4\%) & 2:30 \\
    Elephant & 2 & 181 & 4 (2.2\%) & 15:52 \\
    Zebra & 4 & 336 & 8 (2.4\%) & 16:22 \\
    \cmidrule{1-5}
    \textbf{Total} & \textbf{7} & \textbf{601} & \textbf{14 (2.3\%)} & \textbf{34:44} \\
    \bottomrule
    \end{tabular}%
    }
    \end{minipage}%
    \hfill
    \begin{minipage}[t]{0.02\textwidth}
    \vspace{0pt}\textbf{c}
    \end{minipage}%
    \begin{minipage}[t]{0.37\textwidth}
    \vspace{0pt}
    \resizebox{\linewidth}{!}{%
    \footnotesize
    \begin{tabular}{@{}lccc@{}}
    \toprule
    \multicolumn{4}{@{}l}{\textit{Summary metrics}} \\
    \midrule
     & Rhino & Elephant & Zebra \\
    \midrule
    Tracks & 1 & 1 & 4 \\
    Labelled frames & 84 & 101 & 327 \\
    Overall accuracy & \textbf{0.950} & 0.907 & 0.857 \\
    Subset accuracy & \textbf{0.750} & 0.535 & 0.315 \\
    Mean IoU & \textbf{0.917} & 0.845 & 0.763 \\
    Macro F1 & 0.740 & 0.539 & \textbf{0.661} \\
    \midrule
    \multicolumn{4}{@{}l}{\textit{Per-side F1}} \\
    \midrule
    Front & -- & -- & 0.70 \\
    Back & 1.00 & 0.70 & 0.23 \\
    Left & 0.85 & 1.00 & 0.87 \\
    Right & 0.85 & -- & 0.50 \\
    Top & 1.00 & 1.00 & 1.00 \\
    \bottomrule
    \end{tabular}%
    }
    \end{minipage}

    \vspace{8pt}

    \begin{minipage}[t]{0.02\textwidth}
    \vspace{0pt}\textbf{d}
    \end{minipage}%
    \begin{minipage}[t]{0.97\textwidth}
    \vspace{0pt}
    \includegraphics[width=\textwidth]{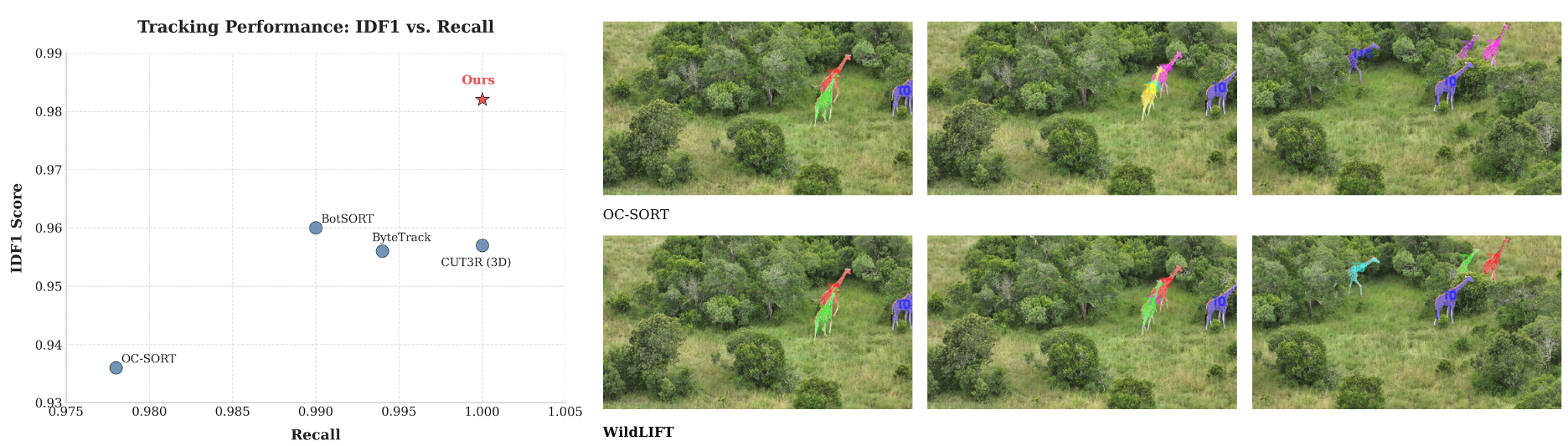}
    \end{minipage}

    \caption{\textbf{Quantitative evaluation of the three WildLIFT modules.}
    \textbf{(a)}~WildLIFT-RT multi-object tracking evaluation across 27 wildlife sequences (6,799 detections, 77 ground-truth individuals).
    Best per metric in \textbf{bold}; second-best \underline{underlined}. Pred~= predicted identity count.
    \textbf{(b)}~WildLIFT-A 3D annotation quality and efficiency.
    \textit{Top:} geometric acceptance indicates frames requiring no correction to position, dimensions, or rotation; semantic flip indicates PCA heading inversion.
    \textit{Bottom:} keyframe interpolation efficiency for the seven tracklets using this mode; total annotation time for all eight tracklets was 41~min.
    \textbf{(c)}~WildLIFT-V viewpoint visibility validation.
    Automated OBB-based viewpoint classification compared against human ground-truth labels across 512 frame$\times$face decisions (5 faces per frame); binary visibility threshold set at 0.1 on the dot-product score (Equation~\ref{eq:visibility}).
    \textbf{(d)}~Tracking performance comparison.
    \textit{Left:} Recall versus IDF$_1$ across all evaluation sequences; 2D trackers sacrifice detection completeness to maintain identity consistency, whereas WildLIFT-RT achieves both perfect recall and the highest IDF$_1$.
    \textit{Right:} Qualitative comparison on a giraffe sequence with prolonged tree occlusion.
    OC-SORT (top) fragments identities when animals re-emerge from vegetation, assigning new track IDs (colour changes); WildLIFT-RT (bottom) maintains consistent identities through occlusion by leveraging 3D position memory via the dormant track pool.}
    \label{fig:evaluation}
\end{figure*}

To assess tracking robustness---particularly in complex multi-animal scenes---we evaluated WildLIFT-RT against a ground truth of 6,799 annotated 3D detections (Fig.~\ref{fig:evaluation}a).
We benchmark our approach against three state-of-the-art 2D trackers---ByteTrack~\cite{zhang2022bytetrack}, BotSORT~\cite{aharonBoTSORTRobustAssociations2022}, and OC-SORT~\cite{caoObservationCentricSORTRethinking2023a}---as well as a 3D ablation baseline using nearest-neighbour association without Kalman filtering (CUT3R~3D).
To isolate the tracking performance, all methods were provided with identical detections from Grounded-SAM.

Across all sequences, WildLIFT-RT achieved perfect recall (1.000) while attaining the highest identity consistency (IDF$_1$~=~0.982), representing a 2.2 percentage-point improvement over the next-best 2D tracker (Fig.~\ref{fig:evaluation}a).
The predicted identity count (82) was closest to the ground truth (77), whereas 2D trackers produced 17--38\% over-counts.

\begin{figure}[t]
    \centering
    \includegraphics[width=0.97\linewidth]{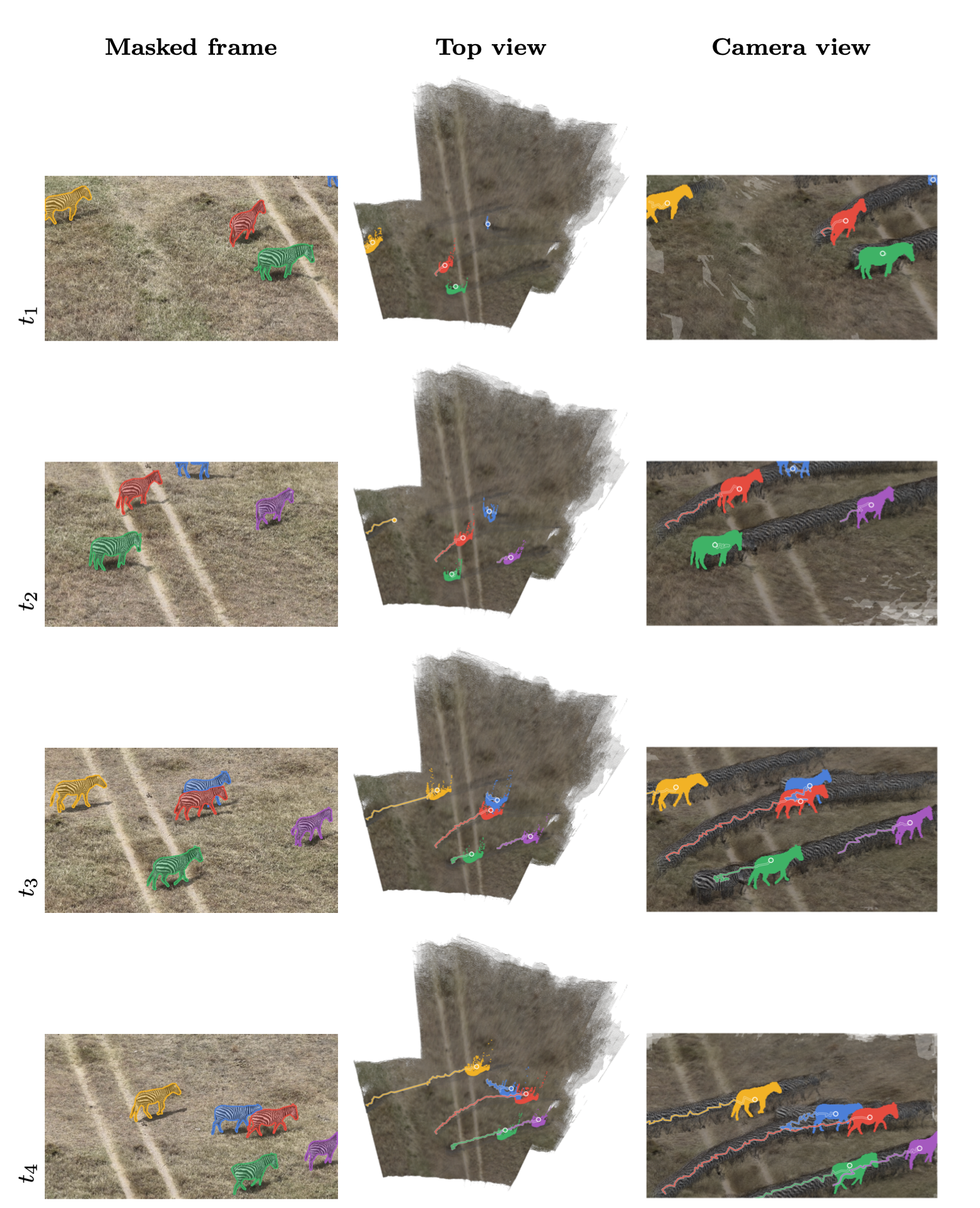}
    \caption{\textbf{3D tracking resolves identity ambiguity in a multi-zebra scene.}
    \textit{Left:} Input frames with instance segmentation masks; colours denote track identity.
    \textit{Centre:} Bird's-eye view of the reconstructed point cloud with per-frame 3D centroids and trajectory traces.
    \textit{Right:} Camera-perspective view of the reconstructed point cloud showing spatial separation along the depth axis.
    The figure illustrates two disambiguation mechanisms.
    First, during the crossover between the red and blue individuals ($t_2$--$t_3$), their 2D masks overlap substantially, yet the 3D centroids remain spatially separated, enabling correct association via the geometric term in Equation~\ref{eq:tracking_cost}.
    Second, the yellow individual (visible at $t_1$, left of frame) exits the field of view at $t_2$ and is absent from the segmentation output.
    The Kalman filter continues to propagate its predicted position in 3D (trajectory trace in centre column); upon re-entering the frame at $t_3$, the detection falls within the re-identification threshold of the dormant track, restoring the original identity rather than initialising a new track.
    Track identity colours are consistent across all panels and frames.}
    \label{fig:tracking}
\end{figure}

Tracking performance varied systematically by species.
Rhino sequences, featuring solitary animals, served as a best-case scenario where all methods achieved perfect scores.
Giraffe sequences, characterised by prolonged vegetation occlusion spanning 30--80 consecutive frames, showed the largest advantage: IDF$_1$ of 0.971 versus 0.922 for BotSORT.
During these extended occlusions, WildLIFT-RT's dormant track pool continued to propagate Kalman velocity estimates, providing predicted positions that remained within the re-identification threshold upon reappearance; the same re-identification mechanism is illustrated for zebras in Fig.~\ref{fig:tracking}, where one individual exits and re-enters the field of view across consecutive frames.
Zebra sequences featured brief crossover events where individuals passed within one body length of each other, resulting in severe 2D mask occlusion (Fig.~\ref{fig:tracking}).
Although the 3D centroids remained spatially separated during these events, enabling correct association, BotSORT achieved the highest IDF$_1$ (0.988 versus 0.980) at the cost of recall; its appearance features likely exploit individual stripe patterns during crossovers, a species-specific advantage unavailable to geometry-only approaches.
An unexpected failure mode emerged for elephants, where BotSORT produced 30 predicted identities for 8 ground-truth individuals (IDF$_1$~=~0.891).
Elephants in these sequences moved slowly ($<$~0.5 body lengths per frame) with gradual posture changes; BotSORT's motion model, calibrated for faster-moving objects in standard benchmarks, likely over-predicted these displacements, resulting in association failures and repeated track fragmentation.
Consequently, deploying standard 2D trackers for diverse megafauna often necessitates species-specific motion model tuning---a form of implicit domain adaptation.
WildLIFT bypasses this requirement entirely; by tracking in reconstructed 3D space, it associates actual geometric displacements rather than relying on brittle 2D motion priors.

The 3D ablation without Kalman filtering achieved perfect recall but lower IDF$_1$ (0.957 versus 0.982), confirming that geometry alone outperforms 2D trackers while the Kalman formulation contributes an additional 2.5 percentage points through velocity-aware prediction and re-identification after occlusion.
Detailed species-specific tracking characteristics are provided in the Supplementary Results.


\subsection*{WildLIFT-A: 3D annotation and refinement}
\label{sec:annotation_results}

Oriented bounding boxes are fitted to each instance-level 3D point cluster via PCA, with optional gimbal telemetry constraints for vertical axis stabilisation (Equation~\ref{eq:ground_normal}).
A browser-based human-in-the-loop tool supports keyframe-based interpolation---LERP for position and dimensions, SLERP~\cite{shoemakeAnimatingRotationQuaternion1985} for orientation---enabling users to annotate entire tracklets by editing a small number of keyframes.
Geometric correction and semantic face labelling are decoupled into independent interpolation modes (Fig.~\ref{fig:interpolation_modes}), addressing PCA eigenvector sign ambiguity without requiring re-editing of spatial fits.
Users assign three primary semantic labels (front, top, left); opposing faces are inferred automatically.
Full details of OBB fitting, interpolation, and the annotation interface are provided in the Online Methods (WildLIFT-A).

We assessed WildLIFT-A on eight tracklets across three species (684 frames), recording whether each automated OBB required geometric correction (Fig.~\ref{fig:evaluation}b).
Across all frames, 93\% required no correction to position, dimensions, or rotation, substantially exceeding the 58.5\% acceptance rate reported by LabelAny3D~\cite{yao2025labelany3d} for single-image 3D labelling (noting that the two pipelines differ in input modality).
This improvement likely reflects temporal consistency from video-based reconstruction and gimbal telemetry constraints that stabilise vertical axis orientation.
Geometric corrections were most frequent for elephants (20\%), where variable trunk positioning altered the point cluster shape sufficiently to skew PCA eigenvectors away from the body's principal axes; in these cases, the adjacent-frame copy tool allowed users to propagate a correct fit from neighbouring frames rather than adjusting from scratch.
Rhino tracklets achieved 100\% acceptance, consistent with their compact, box-like body geometry being well approximated by PCA-fitted OBBs.

Semantic heading flips (PCA eigenvector sign ambiguity inverting the heading by 180\textdegree) occurred in 10\% of frames but were corrected independently of geometry through the decoupled semantic interpolation mode.
Across the seven tracklets using keyframe interpolation, users edited 14 frames to annotate 601 (keyframe ratio 2.3\%), with total annotation time of 41~minutes (3.6~seconds per frame; Fig.~\ref{fig:evaluation}b).
By comparison, manual 3D bounding box annotation on LiDAR point clouds typically requires 60--120~seconds per box~\cite{geigerAreWeReady2012}; WildLIFT-A achieves a roughly 20-fold reduction by amortising human effort across interpolated sequences.

\subsection*{WildLIFT-V: Viewpoint coverage analysis}
\label{sec:viewpoint_results}

\begin{figure}[t]
    \centering
    \includegraphics[width=1\linewidth]{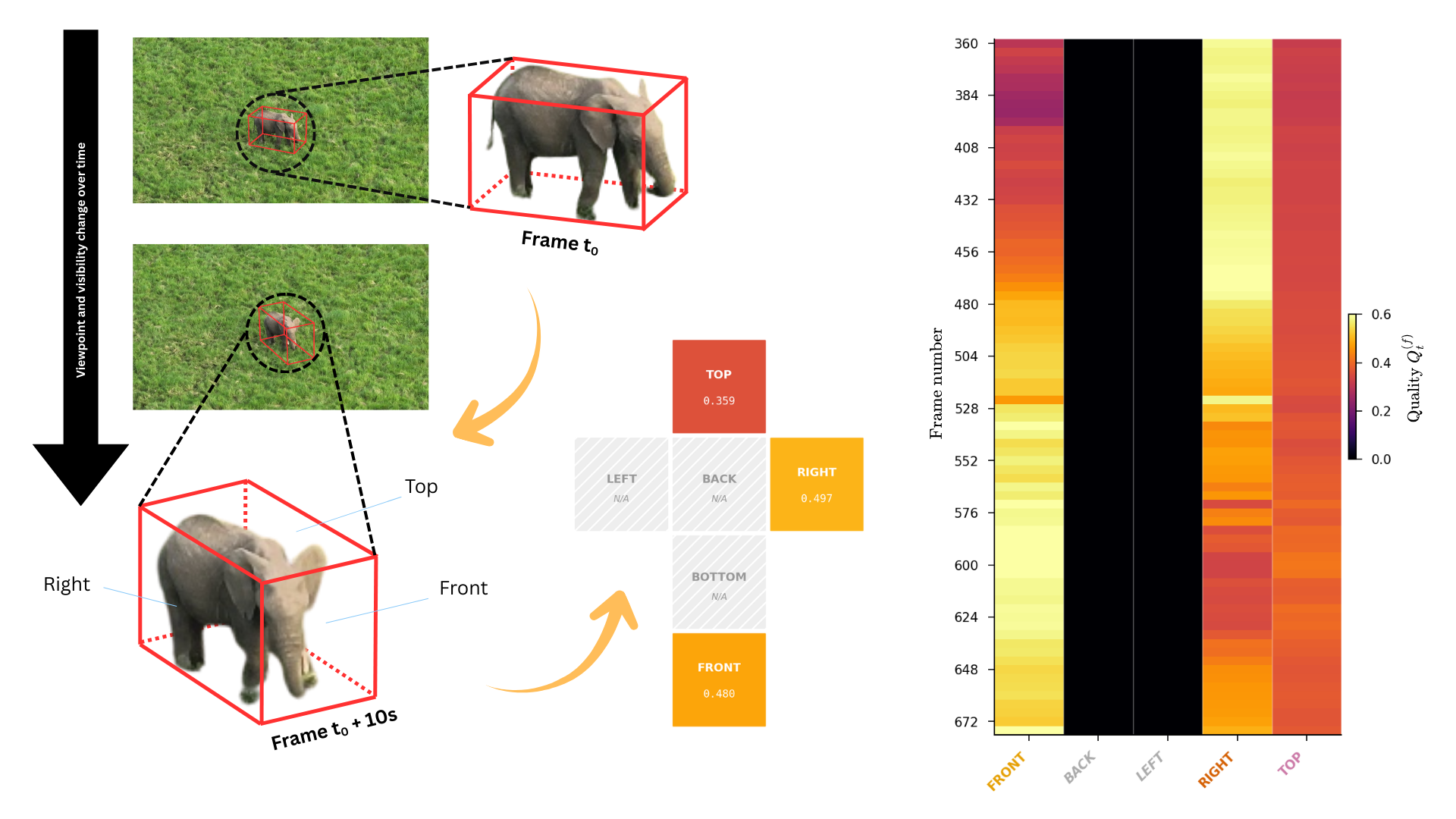}
    \caption{\textbf{WildLIFT-V viewpoint coverage analysis for a single elephant tracklet.}
    \textit{Left:} As the drone repositions over time (frame $t_0$ to $t_0 + 10$\,s), 
    different profiles of the animal become visible. 
    3D oriented bounding boxes recovered in Stage~2 are overlaid on each frame, 
    with semantic face labels (Front, Right, Top) indicated.
    \textit{Centre:} An unfolded bounding box summarising aggregate quality scores 
    (Equation~\ref{eq:quality}) per face across the tracklet. 
    Faces that were never clearly visible (Left, Back, Bottom) are marked N/A, 
    flagging coverage gaps that may require additional flight passes.
    \textit{Right:} Per-frame quality heatmap ($Q_t^{(f)}$) showing how visibility 
    of each profile aspect varies temporally. 
    The Front face transitions from low to high quality mid-sequence as the drone 
    orbits the animal, while the right side becomes less visible. Black columns indicate faces with zero visibility throughout, 
    corresponding to the N/A designations in the unfolded view.
    This temporal decomposition enables programmatic identification of optimal 
    exemplar frames for each viewpoint.}
    \label{fig:viewp_visib}
\end{figure}

Using the semantic face labels from Stage~2, per-frame visibility is determined from the dot product between face normals and the camera viewing direction (Equation~\ref{eq:visibility}).
A continuous quality score (Equation~\ref{eq:quality}) combining viewing angle, projected area, centrality, and foreshortening identifies exemplar frames via temporal non-maximum suppression.
Inter-animal occlusion is quantified through ray--OBB intersection tests, yielding an effective visibility score that combines geometric visibility with occlusion.
Coverage vectors, Shannon entropy-based diversity indices (Equation~\ref{eq:entropy}), and letter grades (A/B/C/F) provide aggregate metadata for batch processing of large video archives.
Full details of the visibility, quality scoring, occlusion analysis, and aggregate metrics are provided in the Online Methods (WildLIFT-V).

WildLIFT-V computes per-frame visibility for five primary viewpoints (front, back, left, right, top) and aggregates these results into coverage matrices (Fig.~\ref{fig:viewp_visib}).
In the representative zebra sequence, all five tracked individuals were observed predominantly from the right flank.
Front, left, and top views were either entirely absent or captured in fewer than three frames across all tracklets.
This distinct coverage bias, driven by the drone's flight path relative to the herd, would require exhaustive frame-by-frame inspection to detect without automated geometric analysis.
Filmstrip visualisations organising exemplar frames by viewpoint are provided in Supplementary Fig.~2.

Mean quality scores across evaluated sequences ranged from 0.53 to 0.67, with higher values corresponding to more orthogonal viewing angles and larger projected areas.
We validated the automated viewpoint classification against human ground-truth labels across 512 frame$\times$face decisions spanning three species (Fig.~\ref{fig:evaluation}c).
Overall accuracy ranged from 0.86 (zebra herds) to 0.95 (rhino), with performance varying by body morphology: compact, box-like bodies produced the cleanest separation between OBB face normals, while elongated bodies in multi-animal scenes presented the greatest challenge.
All classification errors were false positives (over-predicted visibility at shallow angles); recall was approximately 1.0, confirming that the algorithm never missed a truly visible face.
This error asymmetry is practically favourable: false positives at shallow viewing angles produce low quality scores that are unlikely to be selected as exemplar frames, whereas false negatives (missing a truly visible face) would silently discard usable data.

Applying the ray--OBB intersection method to quantify inter-animal occlusion revealed that 15--40\% of geometrically visible frames in multi-animal sequences exhibited partial flank occlusion. 
This finding underscores the critical distinction between geometric visibility (anatomical orientation toward the camera) and effective visibility (an unobstructed line of sight). 
Per-track occlusion statistics and temporal visibility heatmaps are presented in Supplementary Fig.~3.

\section*{Discussion}

Recent advances in feed-forward 3D reconstruction, combined with open-vocabulary segmentation, make it feasible to extract structured 3D representations from the kind of monocular RGB video routinely collected in wildlife monitoring---without requiring specialised sensors, controlled acquisition protocols, or species-specific training.
WildLIFT demonstrates this capability across 27 sequences spanning four morphologically distinct species, three independent data sources, and footage not originally collected for 3D analysis.

The tracking results quantify where 3D geometry provides concrete advantages over 2D approaches.
Off-the-shelf 2D trackers produced 17--38\% identity over-counts across our evaluation sequences; such inflation would require extensive manual verification, offsetting the efficiency gains of automated processing.
WildLIFT-RT's 7\% over-count falls within margins reducible through post-hoc track merging.
Track fragmentation---which distributes an individual's activity budget across pseudo-individuals and dilutes behavioural signals---was reduced by 53\% relative to the next-best 2D method.
These advantages were most pronounced in sequences with prolonged vegetation occlusion, suggesting that 3D tracking offers the greatest benefit in scenarios where extended line-of-sight interruptions exceed the association horizon of 2D methods.
Conversely, for open-habitat species experiencing only brief crossover events, 2D trackers with appearance features may remain competitive, as the zebra results indicate.

The 93\% geometric acceptance rate for automated OBB fitting has implications beyond annotation efficiency.
Generating 3D training data for wildlife presents a circular dependency: supervised 3D detectors require large-scale annotated datasets that do not exist for wildlife, yet producing such annotations without a functional 3D pipeline is prohibitively expensive.
WildLIFT-A partially breaks this cycle by leveraging video-based reconstruction as a pseudo-3D signal, reducing the annotation bottleneck to keyframe-level human verification.
The annotations produced are in KITTI format, directly compatible with existing 3D detection architectures, providing a practical route toward supervised monocular 3D wildlife detection on future, larger-scale datasets.

The viewpoint analysis reveals coverage patterns that are difficult to identify through manual inspection and impractical to quantify systematically across large video collections.
In the evaluated zebra sequence, four of five individuals entirely lacked front, left, and top views---a bias attributable to a single flight path relative to herd orientation.
For bilateral re-identification workflows that require complementary left-flank and right-flank images of the same individual~\cite{schofield_chimpanzee_2019, schneiderPresentFutureApproaches2019}, such undiagnosed gaps render the footage unusable for identity resolution.
WildLIFT-V's coverage vectors and letter grades make such biases explicit and queryable, enabling researchers to filter archives programmatically and to identify which sequences require supplementary flight passes for under-represented orientations.

WildLIFT's use of oriented bounding boxes as a geometric abstraction represents a deliberate trade-off against articulated shape models such as SMAL~\cite{zuffi_3d_2017} and recent benchmarks like Animal3D~\cite{xu_animal3d_2023}.
These approaches recover detailed pose and surface geometry but require species-specific mesh templates, restricting applicability to taxa for which training data exists.
OBBs sacrifice articulated detail for taxonomic generality: the same parameterisation applies equally to rhinos, elephants, zebras, and giraffes without modification, and extends in principle to any taxon detectable by the segmentation model. 
Critically, the six-face structure of the OBB directly enables the face-level visibility analysis in Stage~3 (Fig.~\ref{fig:viewp_visib}): each face serves as a named anatomical proxy, allowing per-frame quality scoring, coverage gap detection, and inter-animal occlusion quantification.
This generality comes at the cost of fine-grained pose information---WildLIFT cannot distinguish a walking from a standing animal---but is sufficient for the spatial reasoning, viewpoint characterisation, and 3D detection training that the framework targets.

The applicability to archival footage is practically significant.
The majority of existing drone-based wildlife video was collected with consumer-grade equipment and cannot be retroactively supplemented with LiDAR or multi-camera rigs.
Our evaluation on data from three independent sources, processed on a consumer-grade GPU (6\,GB VRAM), confirms that the framework operates within the computational and data constraints typical of ecological research groups.

Several limitations should be noted.
WildLIFT's geometric accuracy is bounded by the reconstruction backbone: CUT3R produces internally consistent but scale-ambiguous geometry, and quality degrades under rapid camera motion or severe motion blur.
Our evaluation is restricted to four species of large terrestrial mammals; while the framework's design is not inherently limited to megafauna, extension to birds, marine species, or smaller-bodied taxa will require validation, particularly where body proportions deviate substantially from box-like geometry.
The annotation tool retains a human-in-the-loop requirement; although the 2.3\% keyframe ratio substantially reduces effort, full automation would require learning-based OBB regression---for which WildLIFT's own annotations could provide training data.
Finally, the pipeline currently processes video as a post-hoc analysis step rather than during data acquisition.
CUT3R's online architecture, which processes frames sequentially with constant memory, is in principle compatible with streaming operation, though we have not evaluated latency or throughput in a real-time setting.
The KITTI-format annotations produced by WildLIFT-A provide a standardised training resource compatible with existing 3D detection architectures, potentially enabling supervised monocular 3D wildlife detection as annotated datasets grow.

\begin{figure}[t]
    \centering
    \includegraphics[width=0.85\linewidth]{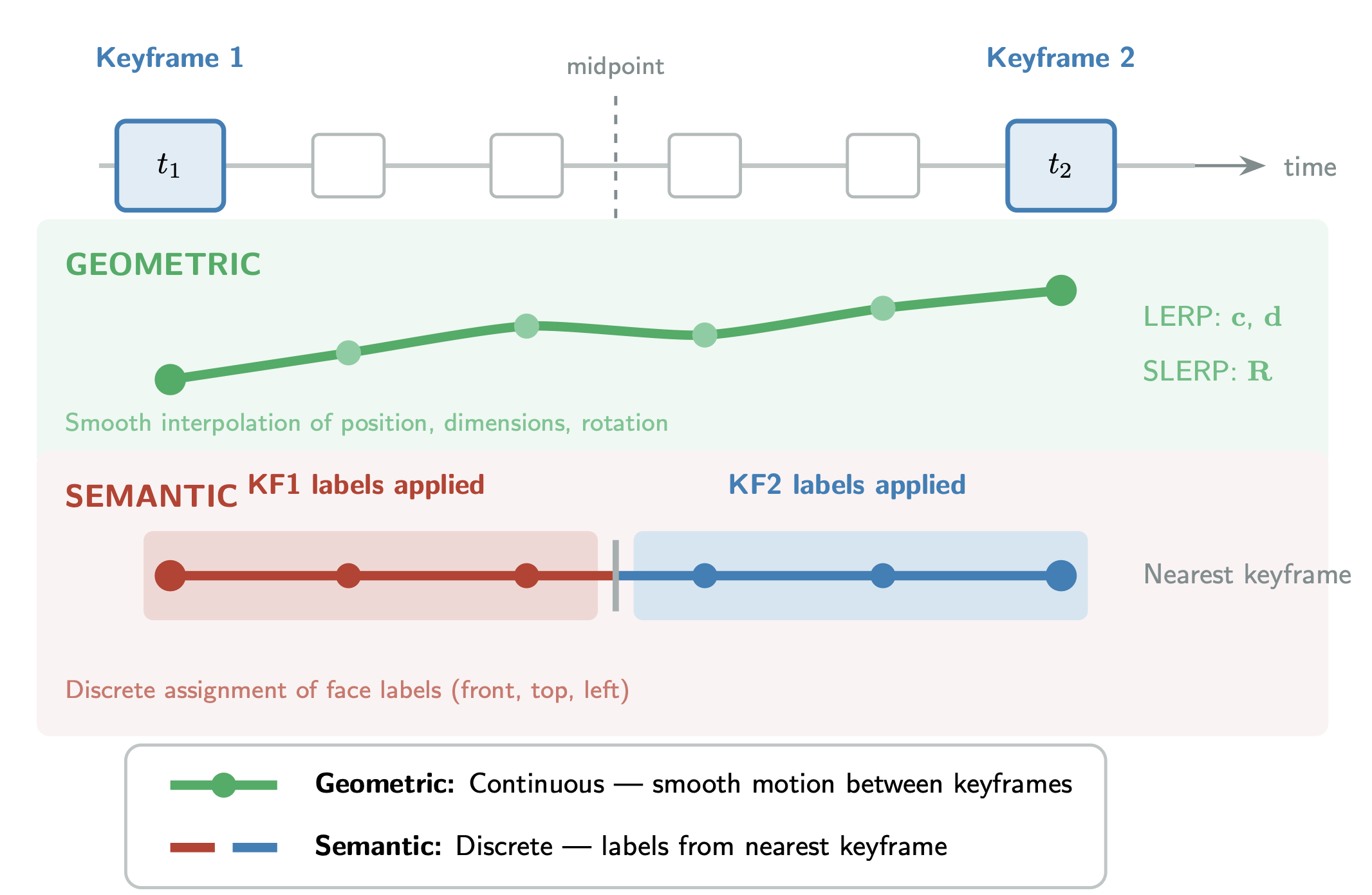}
    \caption{\textbf{Decoupled interpolation modes in WildLIFT-A.}
    \textit{Top:} Geometric interpolation applies continuous LERP
    for position $\mathbf{c}$ and dimensions $\mathbf{d}$
    (Equation~\ref{eq:interpolation}), and
    SLERP~\cite{shoemakeAnimatingRotationQuaternion1985} for
    rotation $\mathbf{R}$, producing smooth motion trajectories
    between keyframes.
    \textit{Bottom:} Semantic interpolation applies discrete
    nearest-keyframe assignment for face labels, accommodating
    cases where an animal's heading changes mid-sequence.
    This decoupling addresses PCA eigenvector sign ambiguity
    (Online Methods, WildLIFT-A): the estimated heading direction
    may be inverted even when the spatial fit is geometrically
    correct. Without decoupling, correcting the heading would
    require re-editing the spatial fit. The two-pass workflow
    (first refine geometry, then update semantic labels) avoids
    this coupling.}
    \label{fig:interpolation_modes}
\end{figure}

\bibliography{sn-bibliography,ref_fin, sn-article-template/ref_fin_nm}

\section*{Online Methods}

\subsection*{WildLIFT-RT: 3D reconstruction and tracking}
\label{sec:methods_rt}

WildLIFT-RT produces a geometric scaffold comprising dense per-frame pointmaps, camera poses, instance-level 3D point clusters for each detected animal, and track identities linking individuals across frames.
This scaffold provides the foundation for subsequent stages.

\paragraph{Feed-forward reconstruction.}
We employ CUT3R~\cite{cut3r}, an online feed-forward transformer for monocular video reconstruction.
CUT3R recovers dense per-frame pointmaps $\mathcal{P}_t \subset \mathbb{R}^3$ and camera poses $\mathbf{T}_t \in SE(3)$ without requiring camera intrinsics or explicit calibration.
The model maintains a persistent state representation across frames, producing temporally consistent geometry, and has been trained on diverse multi-view video data including dynamic scenes.
We selected CUT3R over monocular depth approaches and other feed-forward models after systematic evaluation (Supplementary Table~2, Supplementary Fig.~1): monocular depth networks suffer domain shift on oblique aerial viewpoints~\cite{ranftlRobustMonocularDepth2020}; optimisation-based methods fail when dynamic animals dominate the frame~\cite{liMegaSaMAccurateFast}.
Alternative feed-forward models such as VGGT~\cite{wang2025vggt} explicitly degrade under extreme camera rotations and substantial non-rigid deformation, while also requiring substantially greater computational resources ($>$40 GB VRAM for a 200-frame sequence versus ~5 GB for CUT3R; Supplementary Table 2).

Monocular reconstruction from uncalibrated video produces geometry that is internally consistent but defined in arbitrary units, a property shared by all structure-from-motion methods operating without calibration targets.
WildLIFT operates in a \emph{scale-agnostic} manner, as core functionalities including oriented bounding box fitting, semantic labelling, and viewpoint characterisation rely on relative geometric relationships (proportions, angles, ratios) that remain valid regardless of absolute scale.
For applications requiring metric measurements, scale can be recovered when a reference dimension is available.

\paragraph{Open-vocabulary instance segmentation.}
Concurrently with reconstruction, Grounded-SAM (v2)~\cite{ren_grounded_2024} produces 2D instance segmentation masks $\{M_t^i\}$ from an open-vocabulary text prompt specifying the target species.
This enables deployment to novel taxa without retraining: changing the text query from `zebra' to `elephant' is sufficient to adapt the pipeline to different species.
As open-vocabulary segmentation methods continue to advance, improved models can be substituted without modifying downstream components.

Each 2D instance mask $M_t^i$ is lifted to 3D by selecting the subset of reconstructed points corresponding to masked pixels.
Let $\pi: \mathbb{R}^3 \rightarrow \mathbb{R}^2$ denote projection onto the image plane using the recovered camera parameters.
The 3D point cluster for instance $i$ at frame $t$ is:
\begin{equation}
    \mathcal{Q}_t^i = \{\mathbf{p} \in \mathcal{P}_t : \pi(\mathbf{p}) \in M_t^i\}
\end{equation}
Reconstruction can produce outliers, particularly at object boundaries or in textureless regions.
We apply statistical outlier filtering based on distance from the cluster centroid: points are retained if their distance falls within the interquartile range extended by a factor of 1.5, a standard robust measure that accommodates non-Gaussian distributions common at object boundaries.

\paragraph{Kalman-filtered 3D tracking.}
We associate 3D point clusters across frames to produce coherent per-individual trajectories.
Unlike purely appearance-based 2D trackers, our approach exploits the global consistency of the reconstructed 3D geometry, enabling robust association even under substantial viewpoint changes.

Each active track $\tau$ maintains a Kalman filter~\cite{kalman1960} with a constant-velocity motion model.
The state vector $\mathbf{s}_\tau = [\mathbf{c}_\tau^\top, \dot{\mathbf{c}}_\tau^\top]^\top \in \mathbb{R}^6$ comprises the 3D centroid $\mathbf{c}_\tau$ and its velocity $\dot{\mathbf{c}}_\tau$.
The linear transition model is:
\begin{equation}
    \mathbf{s}_{t+1} = \mathbf{F}\,\mathbf{s}_t + \boldsymbol{\eta}, \quad
    \mathbf{F} = \begin{bmatrix} \mathbf{I}_3 & \Delta t\,\mathbf{I}_3 \\ \mathbf{0} & \mathbf{I}_3 \end{bmatrix}
    \label{eq:kalman_transition}
\end{equation}
where $\Delta t = 1$ (frame interval) and $\boldsymbol{\eta} \sim \mathcal{N}(\mathbf{0}, \mathbf{Q})$ is process noise.
This prediction step propagates each track to its expected position before association, improving robustness when animals move between frames.

For a new detection $d$ with measured centroid $\mathbf{c}_d$ and 2D mask $M_d$, and track $\tau$ with predicted centroid $\hat{\mathbf{c}}_\tau$, the association cost is:
\begin{equation}
    C(d, \tau) = \lambda_{\text{3D}} \cdot \frac{\|\mathbf{c}_d - \hat{\mathbf{c}}_\tau\|_2}{\delta_{\max}} + \lambda_{\text{2D}} \cdot \bigl(1 - \text{IoU}(M_d, M_\tau)\bigr)
    \label{eq:tracking_cost}
\end{equation}
where $\delta_{\max} = 8.0$ (in reconstruction units), $\lambda_{\text{3D}} = 0.7$, and $\lambda_{\text{2D}} = 0.3$.
Associations are constrained to detections sharing the same semantic class.
The 3D term dominates because the reconstructed geometry maintains global consistency, whereas 2D mask overlap can degrade under pose variation.
We apply a \emph{soft gate} on the IoU term: if mask overlap falls below a threshold $\theta_{\text{IoU}} = 0.15$ but the predicted 3D distance remains small (Kalman confidence is high), the match is permitted.
This accommodates appearance changes induced by viewpoint shifts without discarding reliable geometric evidence.
Optimal assignments are computed via the Hungarian algorithm~\cite{kuhnHungarianMethodAssignment1955}.

Tracks are organised into \emph{active} and \emph{dormant} pools to handle occlusion gracefully.
An active track that receives no association for $k_{\text{miss}} = 20$ consecutive frames is demoted to dormant status rather than terminated.
Dormant tracks continue to propagate via the Kalman prediction (equation~\ref{eq:kalman_transition}), preserving their velocity estimate.
Unmatched detections are first compared against active tracks; if no valid assignment exists, a \emph{re-identification} step attempts to match them against dormant tracks using a relaxed distance threshold ($2\,\delta_{\max}$).
Successful re-identification reactivates the track and updates its state, restoring the original identity after occlusion events.
Dormant tracks are permanently removed only after $k_{\text{dormant}} = 100$ frames without reactivation.
This two-tier scheme substantially reduces identity switches in sequences with inter-animal occlusion or temporary detector failures.

Detections that cannot be associated with any existing track initialise new identities.
Upon successful association, the Kalman state is updated with the measured centroid via the standard correction step.
The output of this stage is a set of track identities linking each per-frame detection to a persistent individual, enabling downstream trajectory analysis and identity-consistent bounding box fitting in Stage~2.

\subsection*{WildLIFT-A: 3D annotation with keyframe refinement}
\label{sec:methods_annotate}

WildLIFT-A is a 3D labelling tool for drone-based wildlife data with three complementary capabilities: automated generation of 3D annotations for training monocular detection models, an intuitive interface for verification and refinement, and semantic face labelling to define the anatomical reference frames required for Stage~3 viewpoint analysis.
Using the instance-level 3D point clusters from Stage~1 as a pseudo-3D signal, we generate KITTI-style~\cite{geigerAreWeReady2012} 3D oriented bounding box (OBB) labels through automated fitting followed by human-in-the-loop refinement.

\paragraph{OBB fitting.}
For each frame $t$, we produce a set of OBBs $\mathcal{B}_t = \{b_t^1, \dots, b_t^{N_t}\}$, where each OBB $b = (\mathbf{c}, \mathbf{d}, \mathbf{R})$ comprises a centre $\mathbf{c} \in \mathbb{R}^3$, dimensions $\mathbf{d} = (l, w, h) \in \mathbb{R}^3_+$ representing length, width, and height, and orientation $\mathbf{R} \in SO(3)$.
Each OBB inherits the track identity $\tau$ assigned in Stage~1.
We fit a 3D OBB to each filtered point cluster $\mathcal{Q}_t^i$ using Principal Component Analysis (PCA).
The three eigenvectors of the covariance matrix define the OBB's local coordinate axes, with dimensions determined by the spatial extent of points along each axis.

PCA-based fitting suffers from axis ambiguity: eigenvector signs are arbitrary, so the estimated vertical axis may flip between frames even when the underlying geometry is consistent.
Additionally, oblique viewpoints that capture only partial animal geometry (e.g., a frontal view obscuring body length) can yield poorly oriented boxes.
To mitigate this, we incorporate drone gimbal telemetry to constrain the vertical axis.
Let $\theta_p$ and $\theta_r$ denote the gimbal pitch and roll angles extracted from the drone's telemetry stream (e.g., DJI SRT metadata).
We compute the estimated ground-plane normal as:
\begin{equation}
    \mathbf{n}_g = \begin{bmatrix} \sin\theta_r \\ \cos\theta_p \cos\theta_r \\ -\sin\theta_p \cos\theta_r \end{bmatrix}
    \label{eq:ground_normal}
\end{equation}
We identify the PCA eigenvector most aligned with $\mathbf{n}_g$ (by maximal absolute dot product) and assign it as the OBB's vertical axis.
The remaining two eigenvectors are orthogonalised to span the horizontal plane.
This yields temporally stable orientations without requiring explicit ground-plane segmentation.
When gimbal telemetry is unavailable, the pipeline defaults to unconstrained PCA.

\paragraph{Keyframe-based refinement.}
The core efficiency gain derives from keyframe-based interpolation: rather than correcting each frame independently, users edit OBBs at selected keyframes and the tool propagates corrections to intermediate frames.
For a tracklet spanning $N$ frames, this reduces annotation effort from $N$ independent corrections to editing a small number of keyframe pairs.
Let $\mathcal{B}_{t_1}$ and $\mathcal{B}_{t_2}$ denote user-corrected OBBs at keyframes $t_1 < t_2$.
For each intermediate frame $t \in (t_1, t_2)$, we compute an interpolation parameter $\alpha = (t - t_1)/(t_2 - t_1) \in (0, 1)$.
Centres and dimensions are linearly interpolated:
\begin{equation}
    \mathbf{c}_t = (1-\alpha)\,\mathbf{c}_{t_1} + \alpha\,\mathbf{c}_{t_2}, \quad
    \mathbf{d}_t = (1-\alpha)\,\mathbf{d}_{t_1} + \alpha\,\mathbf{d}_{t_2}
    \label{eq:interpolation}
\end{equation}
Orientations are interpolated on the $SO(3)$ manifold using spherical linear interpolation (SLERP)~\cite{shoemakeAnimatingRotationQuaternion1985} to ensure smooth rotation trajectories that respect the geometry of the rotation group.
Effective keyframe placement depends on motion characteristics: keyframes should bracket segments of consistent motion, as frames exhibiting sudden velocity changes or abrupt direction shifts within an interpolation interval will produce poor intermediate estimates.

\paragraph{Semantic face labelling.}
Downstream viewpoint analysis requires knowledge of which OBB face corresponds to which anatomical aspect.
Users assign three primary semantic labels (front, top, left) to OBB faces via the annotation interface.
The three opposing faces (back, bottom, right) are automatically inferred via surface normal analysis: if a face is labelled `front', the face with maximally anti-parallel outward normal is assigned `back'.
For semantic label propagation specifically, keyframes should be placed where significant orientation changes occur, as labels assigned to box faces must reflect the animal's heading throughout the interval.

Geometric correction and semantic label propagation are decoupled (Fig.~\ref{fig:interpolation_modes}), addressing two practical challenges.
First, PCA eigenvector signs are arbitrary, so the estimated heading direction may be inverted even when the 3D fit is geometrically correct; users can correct heading labels without altering the spatial fit.
Second, if an animal changes orientation during a sequence, users can employ a two-pass workflow: first refine geometry across all frames, then update semantic labels in a separate pass.
The tool supports two interpolation modes: \emph{full interpolation} propagates position, orientation, dimensions, and semantic labels between keyframes; \emph{semantic-only interpolation} propagates face labels while preserving existing geometry, assigning labels from the nearest keyframe based on temporal proximity.

The interactive refinement tool is browser-based, implemented using Viser~\cite{yi2025viser} (\url{https://viser.studio}), a WebGL visualisation framework accessible through any modern web browser without software installation (Supplementary Fig.~4).
Further details of the interface, including ground-snapping functionality, species-proportions constraints, and additional correction modes, are provided in the Supplementary Methods.

\subsection*{WildLIFT-V: viewpoint coverage analysis}
\label{sec:methods_viewpoint}

Using the semantic face labels assigned in Stage~2, WildLIFT-V computes viewpoint coverage statistics that characterise how comprehensively each tracklet captures the animal's appearance from different angles.
Without these labels, notions such as ``front view'' or ``left flank coverage'' would be undefined; the dependency on Stage~2 is therefore structural.
We convert viewpoint coverage to structured metadata, enabling programmatic retrieval of frames capturing specific orientations.

\paragraph{Face visibility.}
For each frame $t$ and semantic face $f \in \{\text{front}, \text{back}, \text{left}, \text{right}, \text{top}\}$, we compute visibility based on the geometric relationship between the face normal and camera viewing direction.
Let $\mathbf{n}_f$ denote the outward unit normal of face $f$, computed from the OBB orientation matrix $\mathbf{R}$.
Let $\mathbf{v}_t = \mathbf{c}_{\text{cam},t} - \mathbf{c}_t$ be the vector from OBB centre to camera centre, where $\mathbf{c}_{\text{cam},t}$ is extracted from the camera pose $\mathbf{T}_t$ recovered in Stage~1.
Face $f$ is visible from the camera if:
\begin{equation}
    \mathbf{n}_f \cdot \hat{\mathbf{v}}_t > 0
    \label{eq:visibility}
\end{equation}
where $\hat{\mathbf{v}}_t = \mathbf{v}_t / \|\mathbf{v}_t\|$ is the unit viewing direction.
The bottom face is excluded from analysis as it is never visible in aerial footage.

\paragraph{Quality scoring.}
Binary visibility indicates \emph{whether} a face is geometrically visible but not \emph{how well} it is captured.
For selecting exemplar frames, we extend visibility to a continuous quality score:
\begin{equation}
    Q_t^{(f)} = \frac{1}{4}\bigl(V_t + A_t + D_t + R_t\bigr)
    \label{eq:quality}
\end{equation}
where $V_t = |\mathbf{n}_f \cdot \hat{\mathbf{v}}_t|$ rewards head-on viewing angles, $A_t$ is the normalised projected face area relative to image dimensions, $D_t$ is a centrality score inversely proportional to distance from image centre, and $R_t = \min(w,h)/\max(w,h)$ penalises foreshortening based on the projected face's aspect ratio.

\paragraph{Temporally diverse exemplar selection.}
We aggregate per-frame scores to identify the best frames for each profile.
For a tracklet of length $T$, we rank frames by $Q_t^{(f)}$ and select exemplars using temporal non-maximum suppression: given a minimum frame separation $\Delta t_{\min}$, we iteratively select the highest-scoring frame that satisfies the temporal constraint with respect to previously selected frames.
This ensures that selected exemplars are temporally distributed rather than clustered around a single high-quality segment.
Alternatively, the interactive mode presents ranked candidates for manual approval, enabling domain experts to apply their judgement in ambiguous cases.

We identify faces that were never clearly visible by detecting semantic labels where the maximum quality score across all frames falls below a threshold.
This ``never-seen'' detection flags gaps in viewpoint coverage that may necessitate additional data collection for tasks requiring comprehensive anatomical documentation.

\paragraph{Inter-animal occlusion.}
In multi-animal scenes, individuals may partially occlude one another, degrading the effective visibility of specific body aspects.
We quantify inter-animal occlusion using ray-OBB intersection tests in 3D, validated against 2D mask overlap.

For each visible face, we cast rays from the camera position through a uniform grid of sample points on the face surface.
Each ray is tested for intersection against all other animals' OBBs using the slab method~\cite{kayRayTracingComplex1986}.
If a ray intersects a nearer OBB before reaching the target face, that sample point is marked as occluded.
Let $\mathcal{S}_f = \{s_1, \ldots, s_n\}$ be sample points on face $f$, and let $\mathcal{B}_{-i}$ denote the set of OBBs for all animals except the target.
For each sample point $s_k$, we define the occlusion indicator:
\begin{equation}
    o_k = \mathbf{1}\bigl[\exists\, b \in \mathcal{B}_{-i} : \text{ray}(\mathbf{c}_{\text{cam}}, s_k) \cap b \neq \emptyset \text{ and } \|b \cap \text{ray}\| < \|s_k - \mathbf{c}_{\text{cam}}\|\bigr]
\end{equation}
The face occlusion percentage is $O_f = (1/n)\sum_k o_k \times 100$.

We combine self-visibility (viewing angle) with inter-animal occlusion into an effective visibility score:
\begin{equation}
    E_f = V_t \cdot (1 - O_f / 100)
\end{equation}
This effective score identifies frames where a face is both geometrically visible and unobstructed by other animals.

As validation, we compute 2D mask overlap using the segmentation masks from Stage~1.
For animals $i$ and $j$ with projected masks $M_i$ and $M_j$, where $i$ is farther from the camera, the overlap percentage is $\text{Area}(M_i \cap M_j) / \text{Area}(M_i)$.
Discrepancies between 3D ray-based and 2D mask-based occlusion can indicate reconstruction errors or mask imprecision.

The analysis produces per-track summaries including: mean visibility across frames, best and worst frames, faces that were consistently occluded, and identification of animals that are ``hard to photograph'' (mean visibility below a configurable threshold).
For each face, we report the frames providing the least-occluded views, enabling researchers to identify optimal frames for viewpoint-sensitive analysis.

\paragraph{Aggregate metrics.}
For batch processing of large video archives, we provide aggregate metrics that summarise viewpoint coverage as single values.
The \emph{coverage vector} $\mathbf{C} \in [0,1]^5$ represents the fraction of frames in which each face is visible:
\begin{equation}
    C_f = \frac{1}{T} \sum_{t=1}^{T} \mathbf{1}[\mathbf{n}_f \cdot \hat{\mathbf{v}}_t > 0]
\end{equation}
This vector can be compared across tracklets to identify individuals with similar or complementary viewpoint coverage.

To quantify whether coverage is balanced across viewpoints or concentrated on a single orientation, we compute a \emph{diversity index} using normalised Shannon entropy:
\begin{equation}
    H = \frac{-\sum_f p_f \log p_f}{\log |F|}
    \label{eq:entropy}
\end{equation}
where $p_f = C_f / \sum_{f'} C_{f'}$ is the normalised visibility proportion and $|F| = 5$.
A diversity index near 1.0 indicates balanced multi-view coverage; values near 0 indicate concentration on a single viewpoint.

\paragraph{Coverage grading.}
For rapid triage, we assign letter grades (A/B/C/F) based on configurable thresholds.
The default grading criterion is the number of viewpoints with at least one high-quality frame: Grade~A requires all five viewpoints covered, Grade~B requires four, Grade~C requires three, and Grade~F indicates fewer than three viewpoints captured.
Tracklets receiving low grades may warrant additional flight passes or may be unsuitable for viewpoint-sensitive analyses.
These thresholds are adjustable to match application-specific requirements.

\bmhead{Acknowledgements}
The WildDrone project \url{https://wilddrone.eu/} has received funding from the European Union's Horizon Europe research and innovation programme under the Marie Sk\l{}odowska-Curie grant agreement no. 101071224. 

\bmhead{Author contributions}
V.S. conceived the framework, designed and implemented the software, curated the evaluation dataset, performed all experiments, and wrote the manuscript.
F.R. and B.R. supervised the project and revised the manuscript. 
B.C. provided ecological domain expertise and revised the manuscript.

\section*{Data availability}
The KABR dataset is publicly available at \url{https://kabrdata.xyz}. 
Some footage from 
Ol Pejeta Conservancy and Bristol Zoological Society is available 
upon reasonable request.

\section*{Code availability}
Code for the WildLIFT framework is available at \url{https://github.com/vandyshukla04/wildlift-framework}. All code required to reproduce the analyses reported in this paper will be deposited in a public repository with a persistent DOI upon publication.

\section*{Competing interests}
The authors declare no competing interests.

\end{document}


\maketitle
\tableofcontents
\clearpage

\section*{Supplementary Figure 1}
\addcontentsline{toc}{section}{Supplementary Figure 1}

\begin{figure}[ht]
    \centering
    \begin{tabularx}{\textwidth}{lXXXXXX}
        \toprule
        \textbf{Species} & \textbf{MegaSaM side} & \textbf{MegaSaM BEV} & \textbf{MegaSaM 3D BBox} & \textbf{CUT3R side} & \textbf{CUT3R BEV} & \textbf{CUT3R 3D BBox}  \\
        \midrule
        
        Rhino & 
        \includegraphics[width=\linewidth]{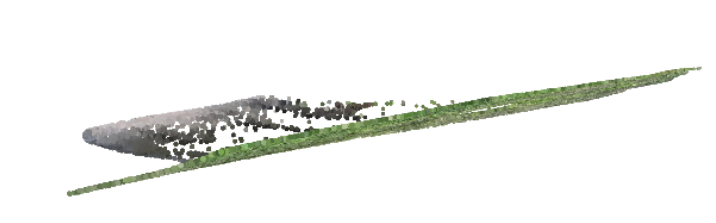} & 
        \includegraphics[width=\linewidth]{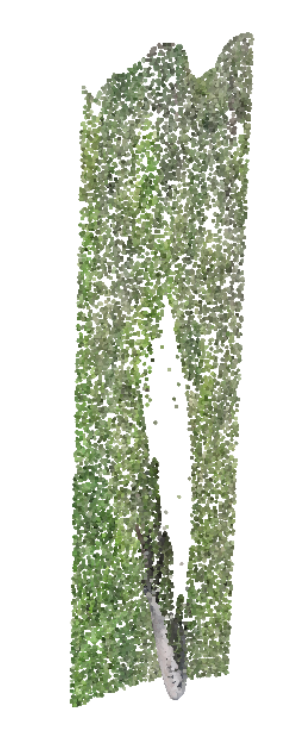} & 
        \includegraphics[width=\linewidth]{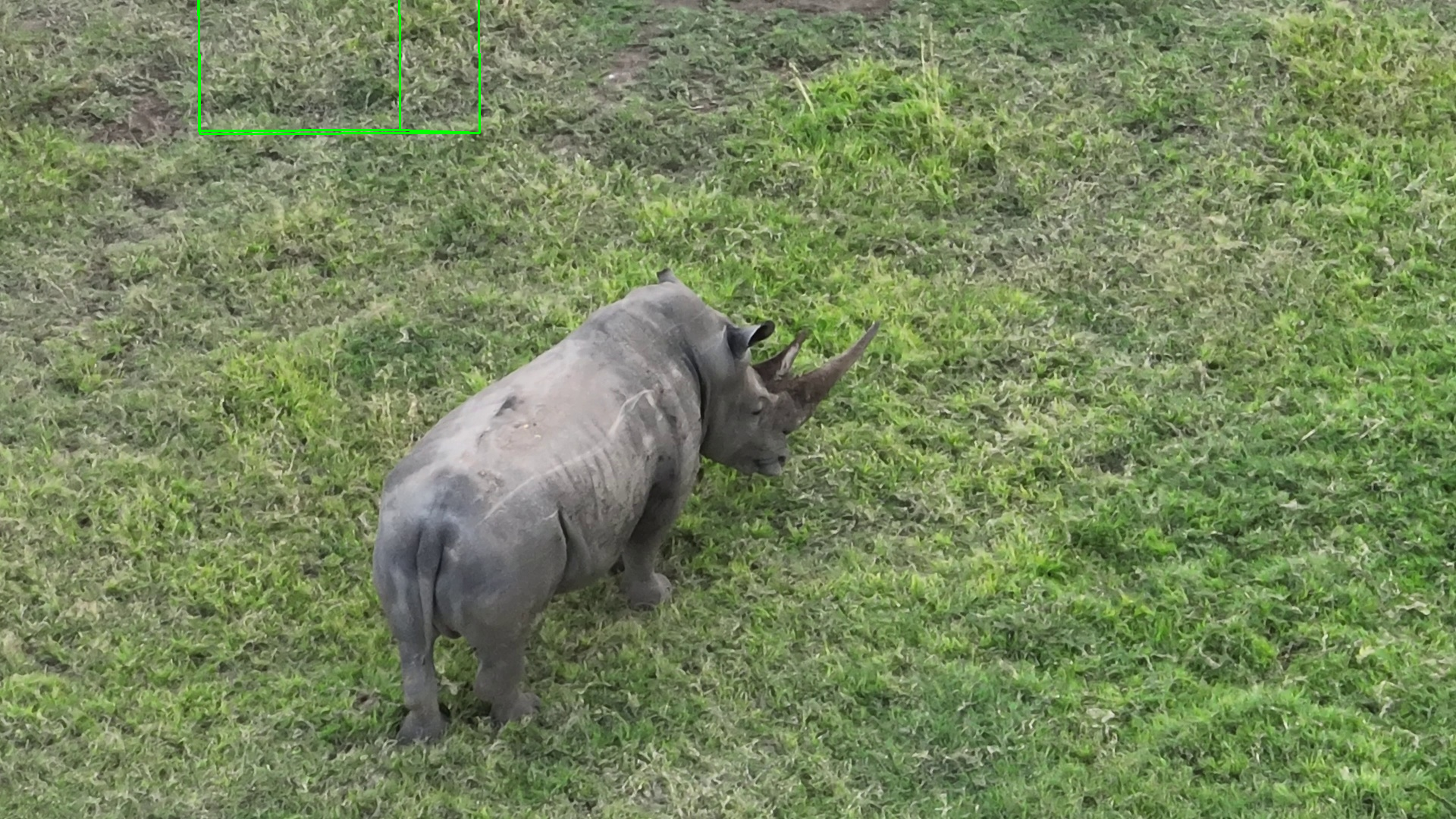} & 
        \includegraphics[width=\linewidth]{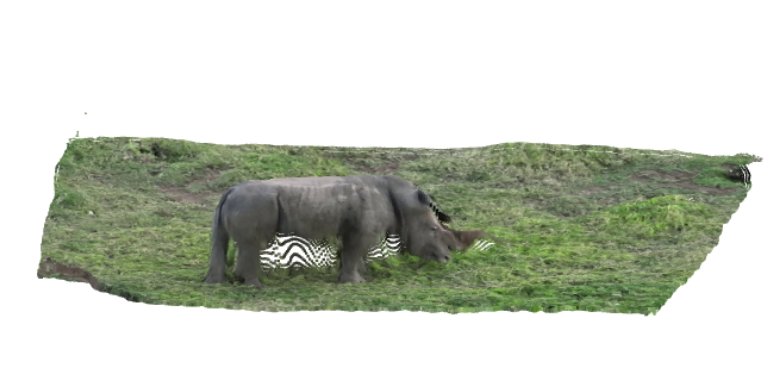} & 
        \includegraphics[width=\linewidth]{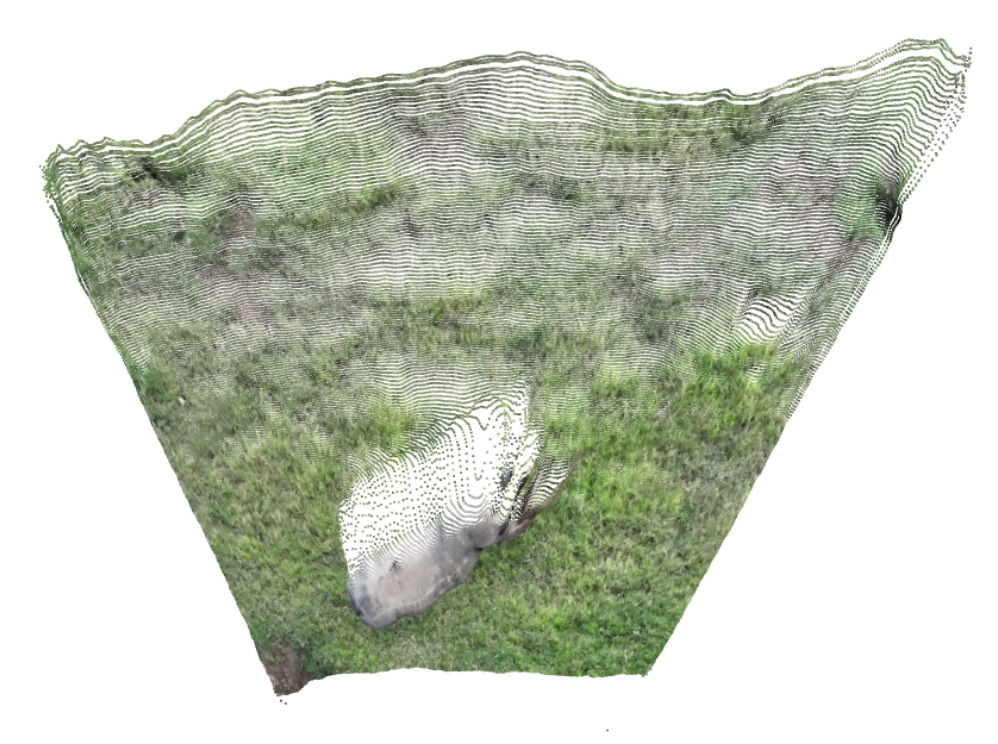} & 
        \includegraphics[width=\linewidth]{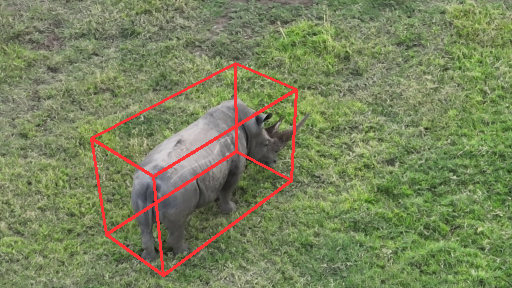} \\
        
        \addlinespace[1em]
        
        Elephant & 
        \includegraphics[width=\linewidth]{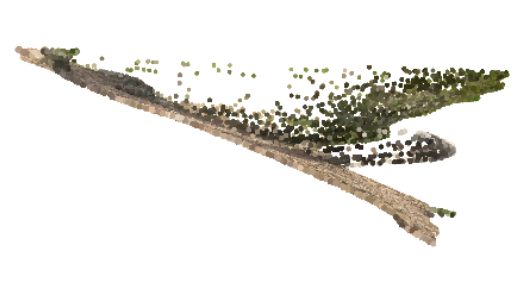} & 
        \includegraphics[width=\linewidth]{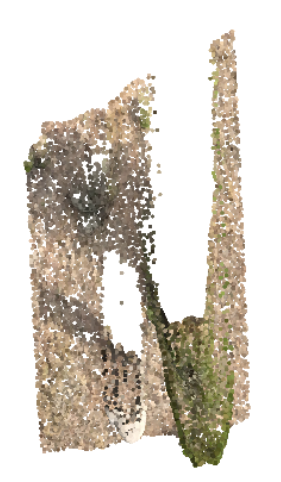} & 
        \includegraphics[width=\linewidth]{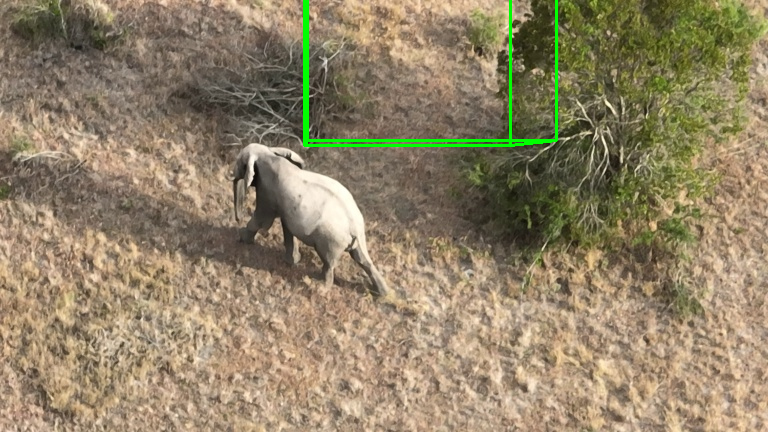} & 
        \includegraphics[width=\linewidth]{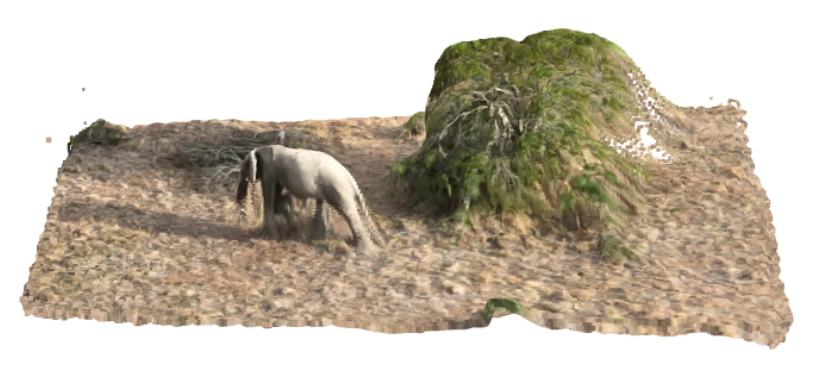} & 
        \includegraphics[width=\linewidth]{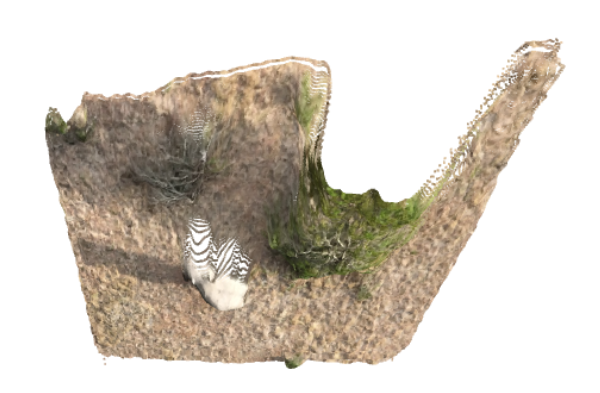} & 
        \includegraphics[width=\linewidth]{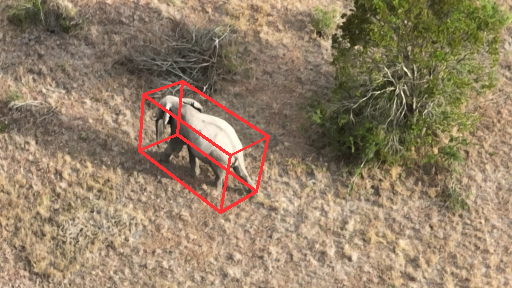} \\
        
        \bottomrule
    \end{tabularx}
    \caption{\textbf{Reconstruction comparison on aerial wildlife footage.}
    MegaSaM~\cite{liMegaSaMAccurateFast}, representative of optimisation-based approaches with mono-depth priors, produces degenerate geometry (collapsed depth planes, incoherent bird's-eye-view structure, inconsistent camera poses evident from misaligned 3D bounding box reprojections) due to domain shift and collinear camera--animal motion.
    CUT3R~\cite{cut3r} learned geometric priors from diverse video data yield plausible 3D structure with coherent spatial relationships between animals and terrain, motivating its adoption as WildLIFT's reconstruction backbone.
    Side views (columns~1 and~4), bird's-eye views (columns~2 and~5), and reprojected 3D bounding boxes overlaid on the input frame (columns~3 and~6) are shown for rhino and elephant sequences.}
    \label{suppfig:megasam_vs_cut3r}
\end{figure}

\clearpage

\section*{Supplementary Figure 2}
\addcontentsline{toc}{section}{Supplementary Figure 2}

\begin{figure}[ht]
    \centering
    \includegraphics[width=\textwidth, page=2]{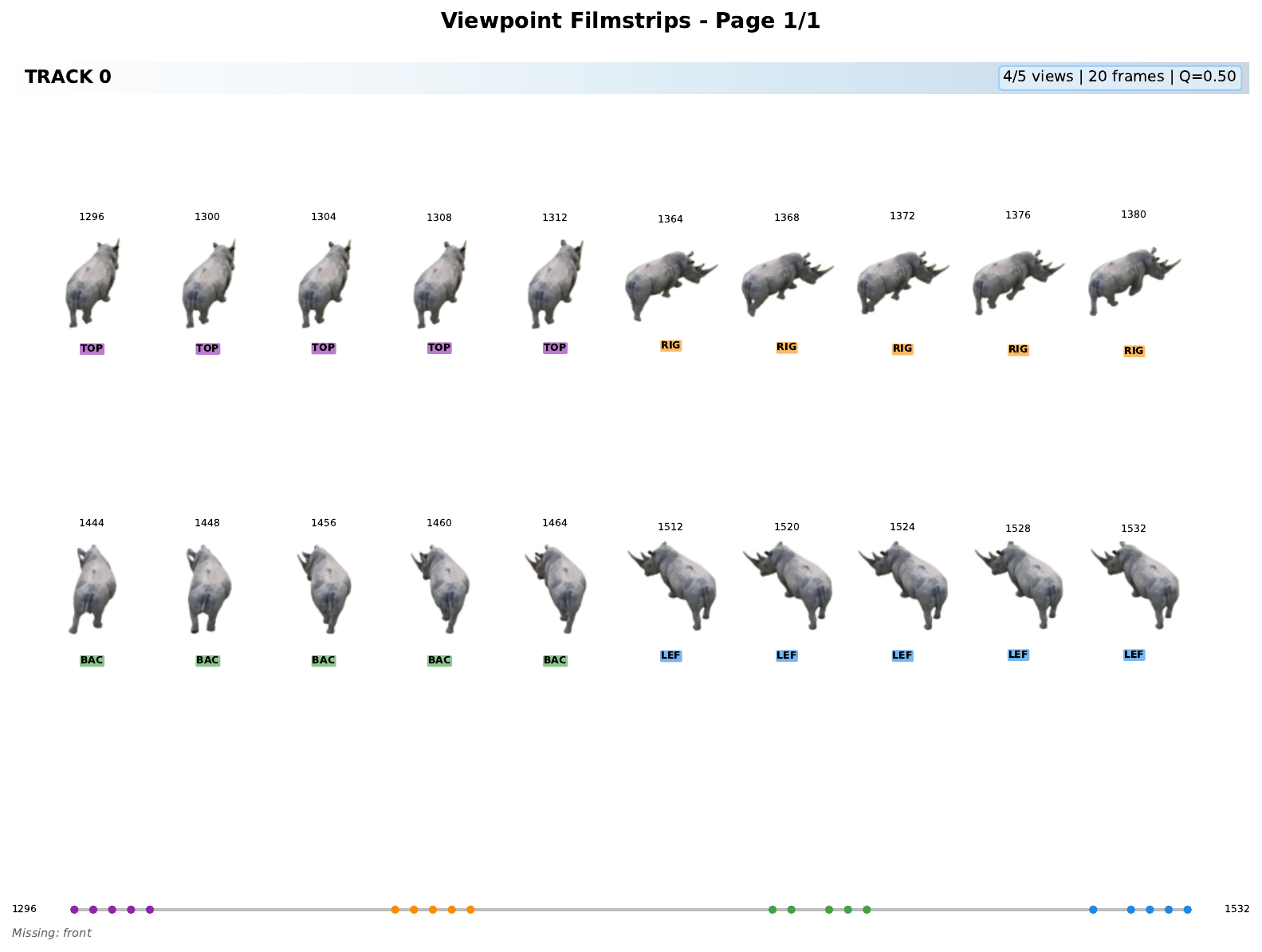}
    \caption{\textbf{Viewpoint filmstrip outputs from WildLIFT-V across three species.}
    For each tracklet, WildLIFT-V selects temporally diverse exemplar frames via non-maximum suppression and organises them by viewpoint (TOP, RIG, BAC, LEF, FRO).
    A timeline at the bottom of each filmstrip colour-codes frames by the dominant visible viewpoint, with missing viewpoints noted.
    \textbf{(a)}~Rhino (rhin-11): 4/5 viewpoints captured; the front view is missing because the drone never orbited to a head-on position.
    (Continued on next pages.)}
    \label{suppfig:filmstrips}
\end{figure}

\clearpage

\begin{figure}[ht]
    \ContinuedFloat
    \centering
    \includegraphics[width=\textwidth, page=2]{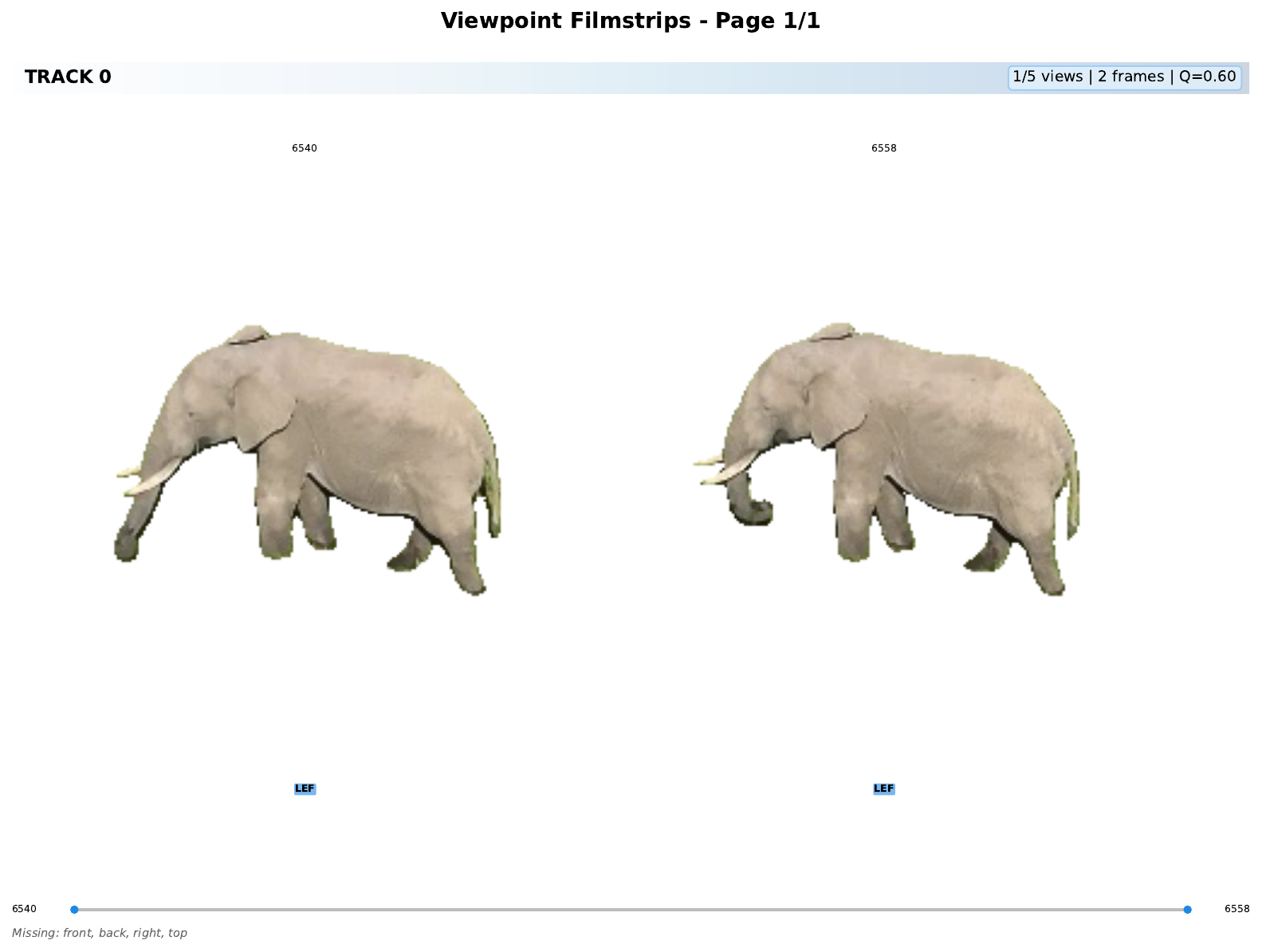}
    \caption[]{\textbf{(Continued.)}
    \textbf{(b)}~Elephant (elep-8\_1): only 1/5 viewpoints captured---the left flank was visible in just 2~frames (mean quality 0.60), while front, back, right, and top views were entirely absent.
    This short tracklet with a near-stationary drone position exemplifies the type of severe coverage gap that WildLIFT-V is designed to flag---a Grade~F assessment indicating that this acquisition is unsuitable for viewpoint-sensitive tasks without additional flight passes.}
\end{figure}

\clearpage

\begin{figure}[ht]
    \ContinuedFloat
    \centering
    \includegraphics[width=\textwidth, page=2]{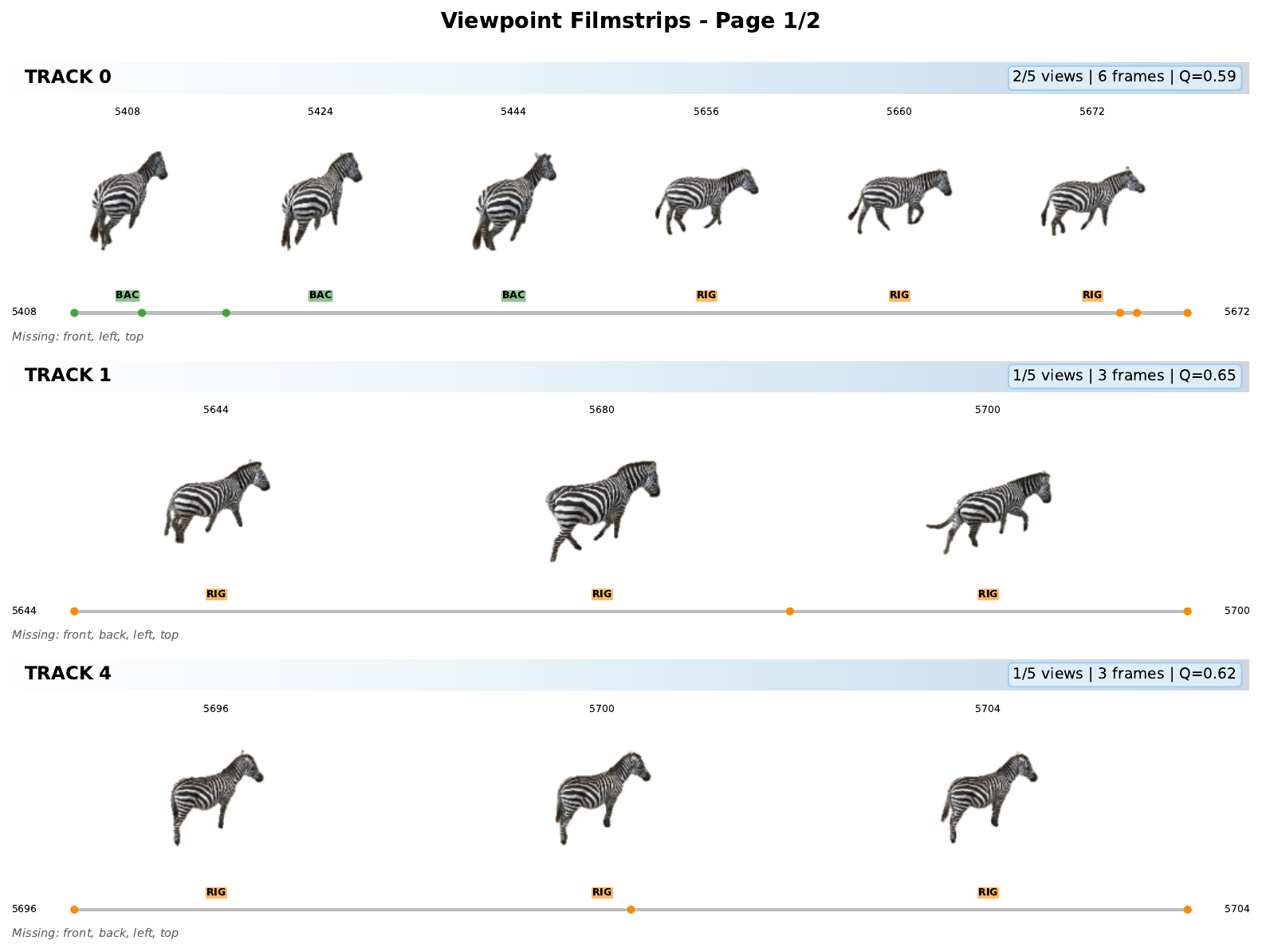}
    \caption[]{\textbf{(Continued.)}
    \textbf{(c)}~Zebra herd (zebr-14\_2), page~1 of~2: multi-track scene showing how individuals in the same group receive different viewpoint coverage depending on their position relative to the drone trajectory.
    (Continued on next page.)}
\end{figure}

\clearpage

\begin{figure}[ht]
    \ContinuedFloat
    \centering
    \includegraphics[width=\textwidth, page=3]{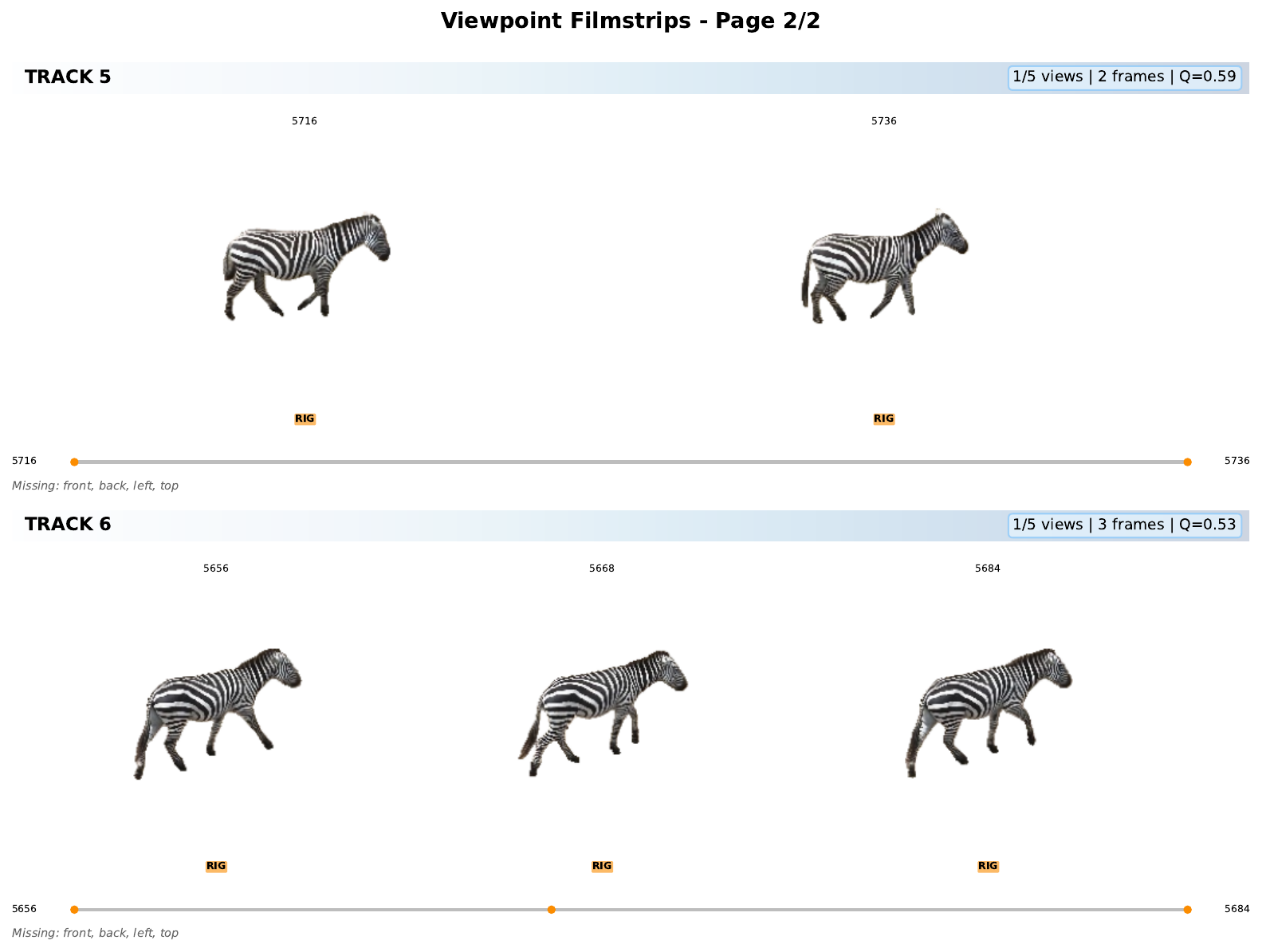}
    \caption[]{\textbf{(Continued.)}
    \textbf{(c)}~Zebra herd (zebr-14\_2), page~2 of~2: remaining tracks from the same herd sequence.
    Across the three species, coverage ranges from 4/5 viewpoints (rhino) to 1/5 (elephant), illustrating the diagnostic value of automated viewpoint analysis.
    These filmstrips are the primary deliverable of WildLIFT-V, providing researchers with an at-a-glance summary of which aspects were captured and at what quality.}
    \label{suppfig:filmstrips_end}
\end{figure}

\clearpage

\section*{Supplementary Figure 3}
\addcontentsline{toc}{section}{Supplementary Figure 3}

\begin{figure}[ht]
    \centering
    \textbf{(a)}\\[0.3em]
    \includegraphics[width=0.3\textwidth]{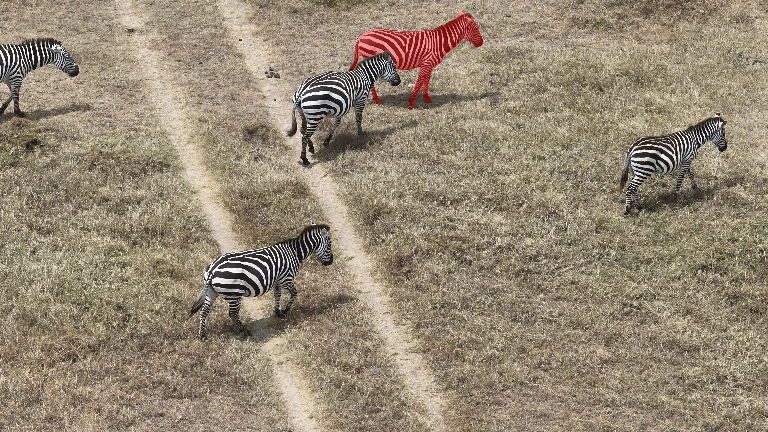}\hfill
    \includegraphics[width=0.3\textwidth]{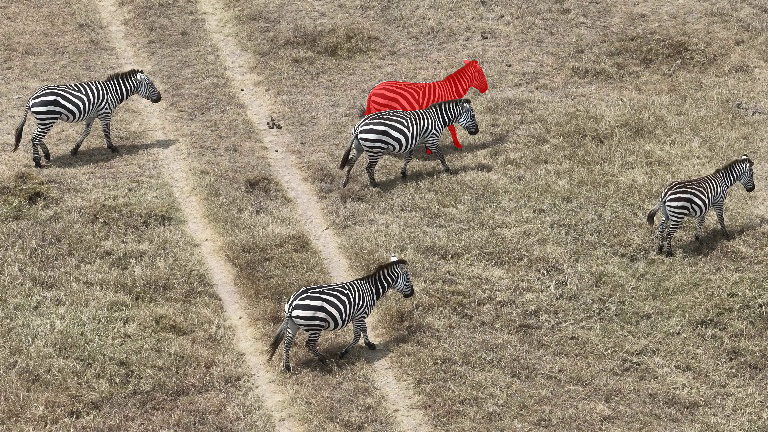}\hfill
    \includegraphics[width=0.3\textwidth]{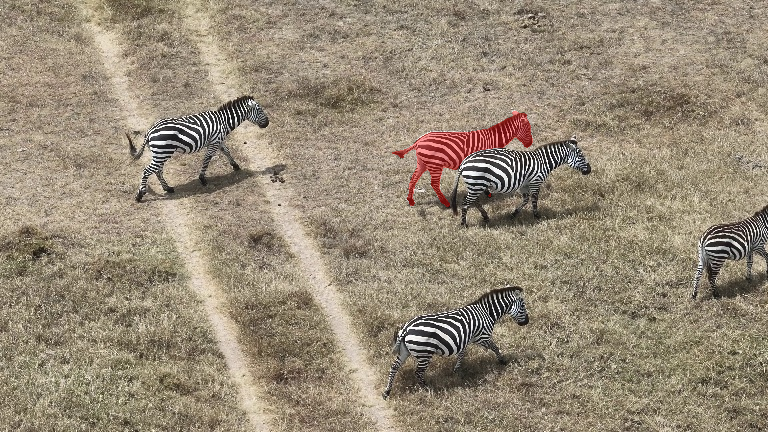}\\[1em]
    \textbf{(b)}\\[0.3em]
    \includegraphics[width=0.5\textwidth, page=1]{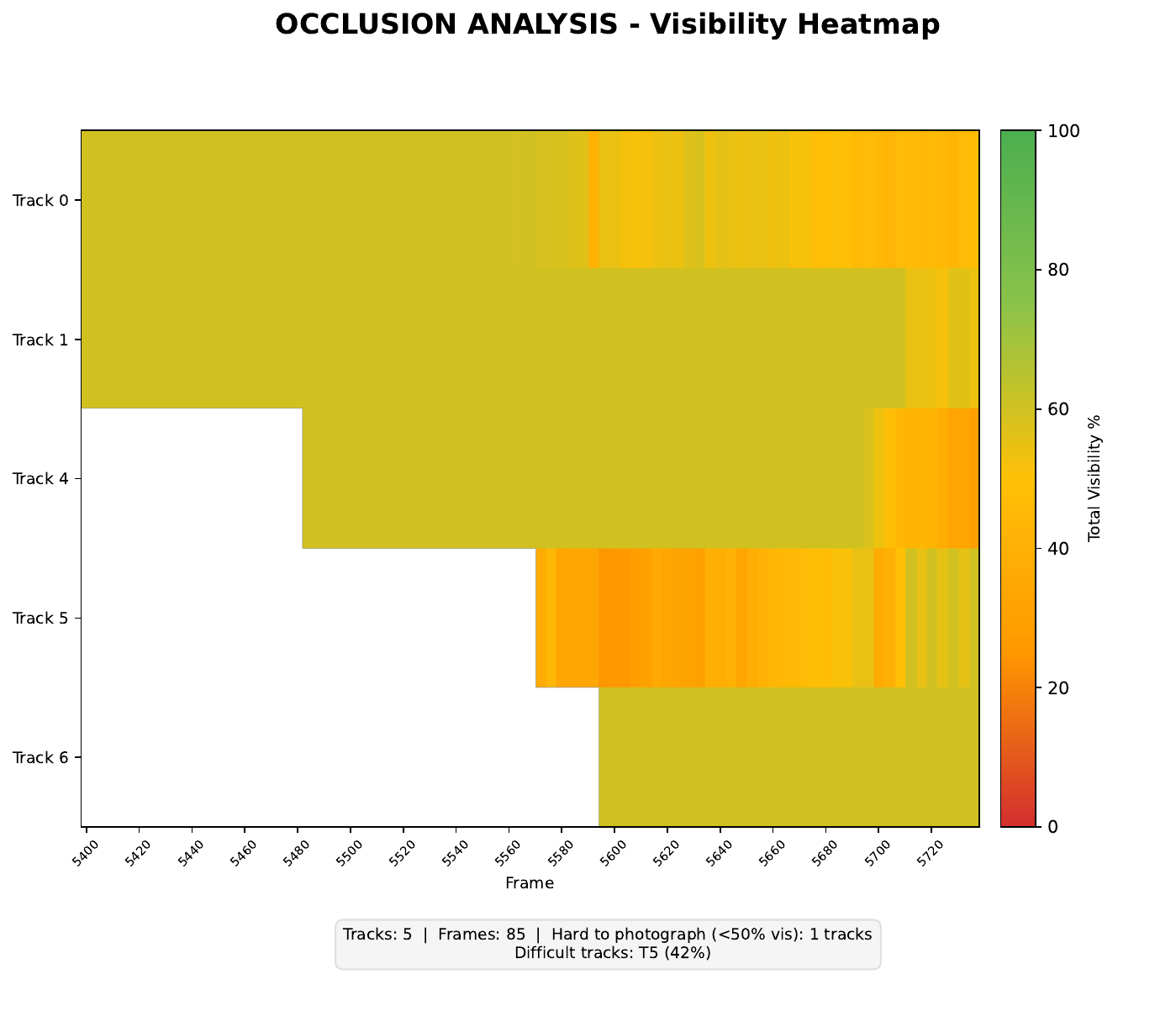}
    \caption{\textbf{Inter-animal occlusion analysis in a zebra herd sequence (zebr-14\_2).}
    \textbf{(a)}~Three frames tracking the occlusion of Track~5 (highlighted in red) over time. 
    Effective visibility fluctuates as the animals move relative to one another and the drone.
    Occlusion is computed via ray--OBB intersection tests from the camera position through the target animal's bounding box faces.
    \textbf{(b)}~Per-track effective visibility heatmap across all 85~frames.
    Each row represents a tracked individual; colour encodes total visibility percentage (red = low, green = high).
    White regions indicate frames before a track was initialised.
    Track~5 is automatically flagged as ``hard to photograph'' (mean visibility 42\%, below the 50\% threshold), while Track~6 maintains consistently high visibility (60\%) because no neighbouring animals occlude its visible faces.
    This heatmap enables researchers to identify which individuals and which temporal segments provide the cleanest data for downstream analysis, without frame-by-frame inspection.}
    \label{suppfig:occlusion}
\end{figure}

\clearpage

\section*{Supplementary Figure 4}
\addcontentsline{toc}{section}{Supplementary Figure 4}

\begin{figure}[ht]
    \centering
    \includegraphics[width=\textwidth]{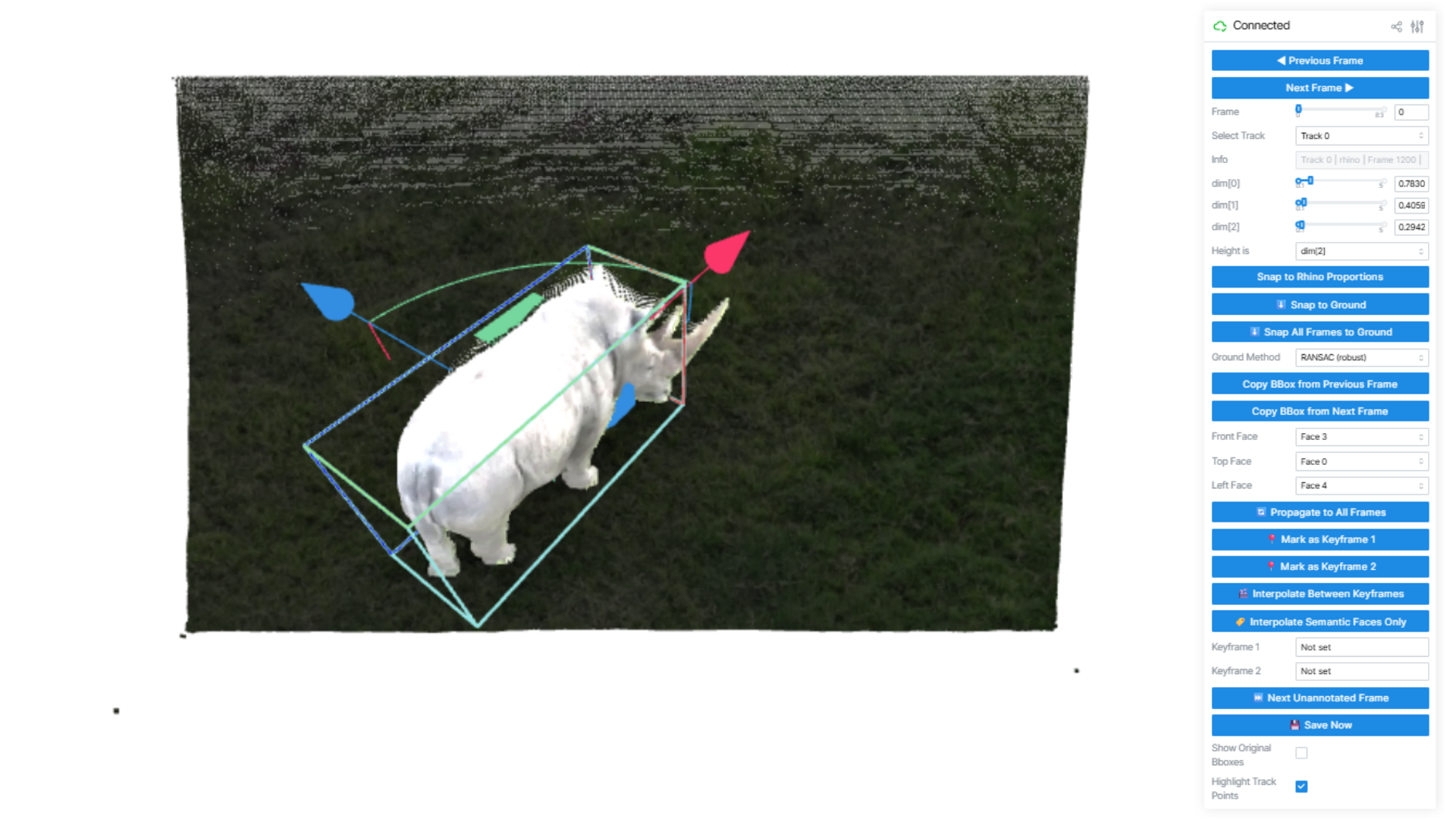}
    \caption{\textbf{WildLIFT-A annotation interface.}
    The browser-based tool displays the reconstructed pointmap with OBB wireframes overlaid.
    Selecting a track activates a 6-DOF transform control for adjusting position and orientation, dimension sliders for box extents along each local axis, and dropdown menus for semantic face assignment.
    Points belonging to the selected animal are highlighted via mask back-projection from the 2D segmentation mask into the 3D pointmap; non-selected points are rendered with reduced opacity, providing immediate visual feedback on whether the OBB accurately encompasses the animal's spatial extent.
    The interface supports frame-by-frame navigation and track selection.
    The tool is implemented using Viser, a WebGL visualisation framework accessible through any modern web browser without software installation.}
    \label{suppfig:annotation_tool}
\end{figure}

\clearpage

\section*{Supplementary Table 1}
\addcontentsline{toc}{section}{Supplementary Table 1}


\begin{table}[ht]
\centering
\caption{\textbf{Evaluation dataset summary.}
Individuals indicates the total number tracked across all clips.
Alt.\ is drone altitude above ground level.
Pitch is the gimbal pitch angle ($-90^{\circ}$ = nadir).
Size is the per-individual segmentation mask area as a percentage of the image.}
\label{supptab:dataset_summary}
\resizebox{\textwidth}{!}{%
\begin{tabular}{@{}llccccccc@{}}
\toprule
Source$^{\dagger}$ & Species & Videos & Individuals & Frames & FPS & Alt.\ (m) & Pitch ($^{\circ}$) & Size (\%) \\
\midrule
\multirow{4}{*}{Ol Pejeta}
    & Rhino     & 8  & 9  & 703 & 3.75--7.5 & 21--85 & $-57$ to $-16$ & 0.5--10.9 \\
    & Zebra     & 5  & 27 & 478 & 7.5--15   & 4--81  & $-46$ to $-14$ & 0.1--2.5 \\
    & Elephant  & 5  & 8  & 539 & 6--10     & 3--79  & $-$24 to $-$1  & 0.1--2.9 \\
    & Giraffe   & 6  & 21 & 584 & 10        & 25--47 & $-23$ to $-16$ & $<$0.1--3.9 \\
\cmidrule{2-9}
    & \textit{Subtotal} & \textit{24} & \textit{65} & \textit{2,304} & \textit{3.75--15} & \textit{3--85} & \textit{$-57$ to $-1$} & \textit{$<$0.1--10.9} \\
\midrule
Bristol Zoo & Zebra & 1 & 2 & 85 & 7.5 & 51 & $-27$ to $-26$ & 0.5--0.9 \\
\midrule
KABR (Mpala) & Zebra & 2 & 10 & 192 & 15 & 16--22 & $-23$ & 0.1--0.6 \\
\midrule
\textbf{Total} & \textbf{4 species} & \textbf{27} & \textbf{77} & \textbf{2,581} & \textbf{3.75--15} & \textbf{3--85} & $\mathbf{-57}$ \textbf{to} $\mathbf{-1}$ & $\boldsymbol{<}$\textbf{0.1--10.9} \\
\bottomrule
\end{tabular}%
}

\vspace{0.5em}
\noindent{\footnotesize $^{\dagger}$Ol Pejeta sequences were captured at effective focal lengths of 24--664\,mm (1--6.7$\times$ digital zoom); Bristol Zoo at a fixed 214\,mm; KABR at a fixed 224\,mm.}
\end{table}

\clearpage

\section*{Supplementary Table 2}
\addcontentsline{toc}{section}{Supplementary Table 2}

\begin{table}[ht]
\centering
\footnotesize
\caption{\textbf{Reconstruction approaches for monocular video of dynamic scenes.}
We evaluated candidate reconstruction methods for WildLIFT along five axes relevant to aerial wildlife footage: robustness to domain shift (training data vs.\ oblique aerial viewpoints), temporal consistency across frames, ability to handle dynamic scene content (moving animals), whether post-hoc global alignment is required, and computational cost.
Peak VRAM was measured on our 200-frame evaluation sequences where possible; CUT3R was processed on an NVIDIA RTX~2060 (6\,GB), while VGGT required a Tesla A40 (48\,GB) on a high-performance computing cluster.
Per-frame methods (marked with $^*$) report single-frame inference cost, which is constant regardless of sequence length.
CUT3R is the only method that satisfies all criteria while remaining accessible on consumer-grade hardware, motivating its adoption as WildLIFT's geometric backbone.
DA = Depth Anything.}
\label{supptab:reconstruction_comparison}
\begin{tabularx}{\textwidth}{@{}l c c X c X@{}}
\toprule
\textbf{Method} & \textbf{Domain} & \textbf{Temporal} & \textbf{Dynamic} & \textbf{Global} & \textbf{Peak VRAM} \\
 & \textbf{robust} & \textbf{consistent} & \textbf{scenes} & \textbf{align.} & \textbf{(200 frames)} \\
\midrule
MiDaS~\cite{ranftlRobustMonocularDepth2020} / DA~\cite{NEURIPS2024_26cfdcd8}
 & No & No & No & N/A & ${\sim}$2\,GB$^*$ \\
\addlinespace
Video DA~\cite{chen2025videodepthanything_cvpr}
 & No & Yes & No & N/A & ${\sim}$3\,GB \\
\addlinespace
Optimisation~\cite{kopf2021robust}
 & No & Yes & Masked only & N/A & ${\sim}$8\,GB \\
\addlinespace
MegaSaM~\cite{liMegaSaMAccurateFast}
 & No & Yes & Masked; fails with collinear motion & Required & ${\sim}$10\,GB \\
\addlinespace
DUSt3R~\cite{Wang_2024_CVPR} / MonST3R~\cite{zhangMONST3RSIMPLEAPPROACH2025}
 & Yes & Yes & Yes & Required & OOM$^{\dagger}$ \\
\addlinespace
VGGT~\cite{wang2025vggt}
 & Yes & Yes & Similar to CUT3R but fails with substantial non-rigid motion & Not needed & $>$40\,GB \\
\addlinespace
\textbf{CUT3R}~\cite{cut3r}
 & \textbf{Yes} & \textbf{Yes} & \textbf{Yes} & \textbf{Not needed} & $\mathbf{\sim}$\textbf{5\,GB} \\
\bottomrule
\end{tabularx}
\vspace{0.5em}

\noindent{\footnotesize $^*$Per-frame inference; VRAM is constant regardless of sequence length. $^{\dagger}$DUSt3R's pairwise formulation requires global alignment and the memory consumption scales quadratically with frame count. One-ref pairing strategy reduces inference pairs but the alignment step remains the bottleneck.}
\end{table}

The key failure modes motivating this selection are detailed below.

\paragraph{Domain shift.}
Monocular depth networks such as MiDaS~\cite{ranftlRobustMonocularDepth2020} and Depth Anything V2~\cite{NEURIPS2024_26cfdcd8} enable zero-shot depth estimation through training on diverse datasets, but training data is predominantly ground-level.
When applied to oblique aerial viewpoints, learned scene priors---horizon position, ground plane orientation, typical object scales---no longer hold.
Video extensions such as Video Depth Anything~\cite{chenVideoDepthAnything2025} enforce temporal smoothness but do not address the underlying domain shift.

\paragraph{Dynamic content masking.}
Per-video optimisation approaches~\cite{kopfRobustConsistentVideo2021b, zhangStructureMotionCasual2022} achieve temporal consistency by jointly refining camera poses and depth, but rely on masking dynamic regions to satisfy bundle adjustment constraints.
MegaSaM~\cite{liMegaSaMAccurateFast} represents the state-of-the-art in casual video reconstruction, combining differentiable bundle adjustment with learned motion masks and uncertainty-aware optimisation.
However, its reliance on mono-depth priors and explicit masking of dynamic content assumes predominantly static backgrounds---an assumption violated by aerial wildlife footage where animals dominate the frame and camera--subject collinear motion (common in drone tracking shots) produces degenerate geometry (Supplementary Fig.~\ref{suppfig:megasam_vs_cut3r}).

\paragraph{Feed-forward alternatives.}
DUSt3R~\cite{wang_dust3r_nodate} pioneered feed-forward pointmap regression from image pairs but requires processing all frame pairs with costly optimisation for full video, with memory scaling quadratically in sequence length---prohibitive for multi-minute wildlife recordings.
MonST3R~\cite{zhangMONST3RSIMPLEAPPROACH2025} extends DUSt3R to dynamic scenes but inherits these scaling constraints.
Both involve optimisation as a post-processing step.
VGGT~\cite{wang2025vggt} processes entire image sets in a single forward pass, eliminating post-hoc fusion, but degradation was observed for some videos under a combination of extreme camera rotations and substantial non-rigid deformation.
However, the main challenge is with VGGT's single-pass architecture, which demands that all frames reside in GPU memory simultaneously: a 200-frame sequence consumed over 40\,GB of VRAM, necessitating a Tesla A40 (48\,GB) on a high-performance computing cluster.
By contrast, CUT3R's recurrent architecture processes frames sequentially with a constant memory footprint of approximately 5\,GB; we processed all 27 evaluation sequences on a single NVIDIA RTX~2060 (6\,GB VRAM), a consumer-grade GPU costing under \$400.
This ${\sim}$8$\times$ reduction in memory requirement is practically significant for ecological research groups operating with constrained computational budgets.
Any4D~\cite{karhadeAny4DUnifiedFeedForward2025} is a concurrent unified feed-forward method but was not publicly available at the time of evaluation.

CUT3R~\cite{cut3r} addresses these limitations through a recurrent architecture maintaining persistent state across frames, producing globally consistent pointmaps without post-hoc alignment while natively handling dynamic scene content.

\clearpage

\section*{Supplementary Video 1}
\addcontentsline{toc}{section}{Supplementary Video 1}
Overview of the WildLIFT pipeline applied to aerial wildlife footage across multiple species. The video is organised by module. WildLIFT-RT: monocular drone video is reconstructed into a dense 3D point cloud with instance-level segmentation masks lifted to 3D; Kalman-filtered tracking produces identity-consistent trajectories through multi-animal scenes. WildLIFT-A: automated PCA-based oriented bounding boxes are fitted to each tracked instance and visualised as reprojected 3D labels on the original frames; the browser-based refinement tool demonstrates 6-DOF adjustment, keyframe interpolation, and semantic face assignment. 
WildLIFT-V: semantic face labels enable per-frame visibility computation, with the quality heatmap, dominant viewpoint timeline, and coverage distribution illustrated for a representative track.
A high-quality version of the video is available for download \href{https://www.dropbox.com/scl/fi/0dv4xuioweb0nkrxyt8tu/WildLift_supplVid.mov?rlkey=hi3usp7y9dbmwg4zeed5jnfsy&dl=0}{here}.

\clearpage
\section*{Supplementary Methods}
\addcontentsline{toc}{section}{Supplementary Methods}

\subsection*{Extended annotation interface description}

The WildLIFT-A interactive refinement tool (Supplementary Fig.~\ref{suppfig:annotation_tool}) provides the following capabilities beyond the core keyframe interpolation described in the Online Methods.

\paragraph{Ground-snapping.}
The interface provides ground-snapping functionality to align OBB bases with the local ground plane.
Three estimation methods are available:
\begin{enumerate}
    \item \textbf{RANSAC-based}~\cite{fischler_random_1981} plane fitting for robust estimation in cluttered scenes, where ground points may be interspersed with vegetation or animal body points.
    \item \textbf{Percentile-based} estimation using the lowest points in a local neighbourhood, suitable when the ground surface is approximately flat and the majority of low-altitude points correspond to terrain.
    \item \textbf{Track-based} estimation using the lowest points of the animal itself (corresponding to feet or hooves), appropriate when the animal is standing on a well-defined surface.
\end{enumerate}
Users select the most appropriate method based on scene characteristics.
Ground-snapping is applied independently of keyframe interpolation and can be toggled on or off per frame.

\paragraph{Species-proportions constraint.}
A species-proportions control enables users to constrain box dimensions to typical anatomical ratios when desired.
This is useful for frames where partial visibility produces PCA-derived dimensions that do not reflect the animal's true proportions (e.g., a frontal view where body length is foreshortened).
The constraint applies fixed ratios between length, width, and height while allowing the overall scale to vary.

\paragraph{Additional correction modes.}
Beyond keyframe interpolation, the interface supports:
\begin{itemize}
    \item \textbf{Adjacent-frame copy}: Copying geometry from an adjacent frame as initialisation for manual correction. This is useful when the automated fit fails on a specific frame but neighbouring frames are correct.
    \item \textbf{Dimension and orientation propagation}: Propagating dimensions and orientation to all frames in a tracklet while preserving per-frame centre positions. This is useful for sequences where an animal remains approximately stationary but the PCA-derived orientation fluctuates due to partial visibility changes.
\end{itemize}

\paragraph{Mask back-projection.}
To aid spatial reasoning during manual correction, the tool highlights points belonging to the selected animal by back-projecting its 2D segmentation mask into the 3D pointmap.
Non-selected points are rendered with reduced opacity, providing immediate visual feedback on whether the OBB accurately encompasses the animal's spatial extent.

\subsection*{Ray--OBB occlusion sampling details}

For inter-animal occlusion analysis (Online Methods), we cast rays from the camera position through a uniform grid of sample points on each visible face.
The sampling grid uses $8 \times 8 = 64$ points per face by default, uniformly distributed in the face's local 2D coordinate system.
Each ray is parameterised as $\mathbf{r}(t) = \mathbf{c}_{\text{cam}} + t(\mathbf{s}_k - \mathbf{c}_{\text{cam}})$, where $\mathbf{s}_k$ is the sample point and $t \in [0, 1]$.

Intersection testing uses the slab method~\cite{kayRayTracingComplex1986}, which tests a ray against each pair of parallel planes defining the OBB.
A ray intersects the OBB if and only if the intersection of all three slab intervals is non-empty.
For each sample point, we test the ray against all other animals' OBBs ($\mathcal{B}_{-i}$) and record the nearest intersection distance.
If this distance is less than the distance to the target sample point, the sample is marked as occluded.

The $8 \times 8$ default grid provides a balance between spatial resolution and computational cost.
For the evaluation sequences (up to 7 animals per frame, 5 visible faces per animal), occlusion computation takes less than 0.1 seconds per frame on a standard CPU, making it practical for batch processing.

\subsection*{Expanded Kalman filter and association details}

\paragraph{Process and measurement noise.}
The process noise covariance $\mathbf{Q}$ is set as a diagonal matrix with position variance $\sigma_q^2 = 0.5$ and velocity variance $\sigma_v^2 = 0.1$ (in reconstruction units per frame).
The measurement noise covariance $\mathbf{R}_{\text{meas}}$ uses a diagonal matrix with $\sigma_m^2 = 1.0$.
These values were determined empirically across the evaluation sequences and kept constant across all species.

\paragraph{Soft gate details.}
The soft gate mechanism relaxes the IoU constraint when 3D geometric evidence is strong.
Specifically, if $\text{IoU}(M_d, M_\tau) < \theta_{\text{IoU}} = 0.15$ but the normalised 3D distance $\|\mathbf{c}_d - \hat{\mathbf{c}}_\tau\|_2 / \delta_{\max} < 0.3$, the match is still permitted.
This addresses a common scenario in aerial footage: when the drone changes viewpoint rapidly, the 2D mask appearance of an animal can change substantially (e.g., transitioning from side to top view), causing IoU to drop even though the 3D position is well predicted.
Without the soft gate, these transitions would cause track fragmentation.

\paragraph{Re-identification threshold.}
Dormant track re-identification uses a relaxed distance threshold of $2 \times \delta_{\max} = 16.0$ reconstruction units.
This larger threshold accommodates the accumulated prediction drift during occlusion periods, where the Kalman filter propagates without measurement updates.
The re-identification step first checks class consistency (the detection must match the dormant track's species), then evaluates the distance criterion.
If multiple dormant tracks are within threshold, the nearest is selected.

\clearpage

\section*{Supplementary Results}
\addcontentsline{toc}{section}{Supplementary Results}

\subsection*{Viewpoint visibility validation details}

The viewpoint visibility evaluation summarised in Fig.~2c of the main text was conducted across 512 frame$\times$face decisions (5 faces per frame) spanning three species.
The binary visibility threshold was set at 0.1 on the dot-product score (Equation~6 in the main text).

All classification errors were false positives: recall was approximately 1.0 across all sides and species, meaning the algorithm never missed a truly visible face but over-predicted visibility at shallow viewing angles.
AUC values ranged from 0.83 to 1.00 where both positive and negative ground-truth cases existed, confirming that the continuous visibility scores rank correctly and that errors reflect threshold calibration rather than scoring quality.
The exception was the elephant back face (AUC = 0.53), where the animal's rounded body prevents discriminative dot-product scores between adjacent faces.

Performance varied systematically with body morphology.
Compact, box-like bodies (rhino) produced the cleanest score separation between faces (accuracy 0.95), as the OBB geometry closely approximates the animal's true shape.
Rounded bodies (elephant) reduced discrimination between adjacent faces (accuracy 0.91).
Elongated bodies in multi-animal scenes (zebra herd) presented the greatest challenge (accuracy 0.86), combining the geometric difficulty of high-aspect-ratio OBBs with the additional complexity of inter-animal occlusion.
Geometric and occlusion-adjusted visibility scores differed by less than 0.01 across all metrics in these evaluation sequences, indicating that inter-animal occlusion was not the primary source of classification error.

\subsection*{Annotation efficiency details}

Across the eight evaluated tracklets, annotation characteristics varied by species and scene complexity.

All keyframe interpolation was semantic-only: heading labels (front/back assignment) were propagated between two keyframes per tracklet (start and end), with a mean interpolation span of 85 frames between edits.
Geometric corrections (position, dimensions, rotation) were applied frame-by-frame where needed, using the adjacent-frame copy and ground-snapping tools.
One rhino tracklet required no geometric correction; only per-frame semantic labels were assigned, taking 5 minutes 50 seconds.

Total annotation time for all eight tracklets was 41 minutes, yielding 3.6 seconds per frame on average.
Elephant sequences required the most time (7--9 minutes per tracklet) due to ground-plane corrections necessitated by telemetry gaps and variable trunk positioning.
For one such tracklet, ground-snapping was applied to all frames to compensate for missing gimbal data, and a small number of frames required manual geometry correction propagated via the adjacent-frame copy tool.
Another elephant tracklet exhibited PCA axis flips in a contiguous block of frames (41--67), which were corrected by copying the bounding box from neighbouring frames and adjusting rotation.
Zebra tracklets in the multi-animal scene averaged 4 minutes each despite inter-animal occlusion, because the consistent herd motion produced smooth semantic interpolation trajectories; one zebra track required species-proportions constraints where a foreshortened camera angle compressed the apparent body length.

\subsection*{Per-track occlusion statistics}
\label{app:occlusion}

In the zebra herd evaluation sequence with 5 tracks, inter-animal occlusion affected 4 tracked individuals to varying degrees (Supplementary Fig.~\ref{suppfig:occlusion}).
Mean effective visibility ranged from 42\% (Track~5, the most occluded) to 60\% (Track~6, effectively unoccluded).
Track~5 was automatically flagged with mean visibility below the 50\% threshold, with Track~0 identified as the primary occluder.
Effective visibility for Track~5 fluctuated between 20\% and 47\% across frames, reflecting the dynamic spatial relationship between herd members as they moved relative to one another and the drone (Supplementary Fig.~\ref{suppfig:occlusion}a).

The distinction between geometric visibility (face oriented toward camera) and effective visibility (face oriented toward camera and unobstructed) has practical implications for frame selection.
For applications requiring clean profile images, such as individual re-identification or morphometric measurement, selecting frames based solely on geometric visibility may yield images where the target animal is partially blocked by a neighbouring individual.
The effective visibility score $E_f = V_t \cdot (1 - O_f / 100)$ addresses this by down-weighting frames with inter-animal occlusion, even when the viewing angle is favourable.

Per-track summary outputs include: mean effective visibility across frames for each face, identification of the best and worst frames per face, faces that were consistently occluded, and flagging of animals that undergo occlusion.

\subsection*{Filmstrip exemplar visualisations}
\label{app:filmstrip}

For each tracklet, WildLIFT-V produces filmstrip visualisations that organise selected exemplar frames by viewpoint (front, back, left, right, top), providing a rapid visual summary of viewpoint coverage without reviewing the full video.
Supplementary Figure~\ref{suppfig:filmstrips} shows representative filmstrips for one tracklet per species across three pages.
Exemplar selection uses temporal non-maximum suppression with a minimum frame separation $\Delta t_{\min}$ to ensure temporal diversity.

The rhino filmstrip (Supplementary Fig.~\ref{suppfig:filmstrips}a) captured 4 of 5 viewpoints; the front view was missing because the drone never orbited to a head-on position.
The zebra herd filmstrip (Supplementary Fig.~\ref{suppfig:filmstrips}c) demonstrates how individuals in the same group receive different viewpoint coverage depending on their position relative to the drone trajectory.
This case study illustrates how WildLIFT-V can surface acquisition-specific biases for a single video; the patterns observed reflect the specific flight path and herd configuration rather than general trends across the video collection.
Different flight paths, herd orientations, and species would yield different coverage profiles; the value of the module lies in making such patterns explicit and queryable rather than requiring manual frame-by-frame inspection.

\subsection*{Species-specific tracking characteristics}

This section expands on the species-level tracking results reported in the main text, providing mechanistic explanations for the observed performance patterns.

\paragraph{Giraffes.}
The 4.9 percentage-point IDF1 advantage over BotSORT~\cite{aharonBoTSORTRobustAssociations2022} reported in the main text was driven by prolonged vegetation occlusion events spanning 30--80 consecutive frames---substantially longer than the $k_{\text{miss}} = 20$ threshold for dormant-pool demotion.
The two-tier track management was critical in these sequences: without it, re-emerging animals would be assigned new identities.
During occlusion, the Kalman velocity estimate continued to propagate, providing a predicted position that remained within the relaxed re-identification threshold ($2 \times \delta_{\max}$) upon reappearance.

\paragraph{Zebras.}
During crossover events, individuals passed within one body length of each other, causing 2D mask IoU between crossing animals to momentarily exceed the IoU of the correct track association.
The 3D geometric term ($\lambda_{\text{3D}} = 0.7$) mitigated this: even when 2D masks overlapped substantially, the 3D centroids remained spatially separated because animals occupy different depths relative to the camera.
BotSORT's higher IDF1 on zebra sequences (0.988 versus 0.980) likely reflects its appearance features distinguishing individual stripe patterns during crossovers---a species-specific advantage unavailable to geometry-only approaches.

\paragraph{Elephants.}
The BotSORT failure (IDF1 = 0.891, 30 predicted identities for 8 ground-truth individuals) warrants detailed examination.
Elephants in the evaluation sequences moved slowly ($<$~0.5 body lengths per frame) with gradual posture changes.
BotSORT's motion model, calibrated for faster-moving objects in benchmark datasets, likely predicted larger displacements than observed, causing associations to fail and new tracks to be initialised repeatedly.
This finding carries practical implications: off-the-shelf 2D trackers may require species-specific motion model tuning for megafauna, a form of implicit domain adaptation that WildLIFT's 3D approach avoids because the Kalman filter operates on actual reconstructed positions rather than learned motion priors.

\paragraph{Rhinos.}
All methods achieved perfect scores on rhino sequences (solitary or well-separated individuals), confirming that the additional computational cost of 3D reconstruction does not introduce spurious errors when tracking is unambiguous.
These sequences served as a ceiling condition for validation.

\clearpage

\section*{Supplementary Note 1: Extended related work}
\addcontentsline{toc}{section}{Supplementary Note 1}

This note expands on the related work summarised in the main text, providing additional context on prior approaches to 3D animal analysis and open-vocabulary segmentation that informed WildLIFT's design.

\subsection*{Prior approaches to 3D analysis in ecology}

The main text positions WildLIFT relative to monocular depth estimation and feed-forward reconstruction.
Here we situate the framework within the broader landscape of 3D animal analysis, which spans several complementary paradigms.

\paragraph{Parametric shape models.}
Foundational parametric models such as SMAL~\cite{zuffi_3d_2017} and SMALST~\cite{zuffi_three-d_2019} recover articulated 3D shape from single images by fitting a learned deformable mesh to 2D observations.
Recent benchmarks including Animal3D~\cite{xu_animal3d_2023} and Animal-in-Motion~\cite{Zhao2025NeurIPS} extend articulated shape extraction to videos, while generative approaches such as RAW~\cite{kulits2025raw} reconstruct broader environmental contexts.
These methods produce detailed articulated geometry but require species-specific shape priors (training meshes per taxon), which limits their applicability to the four-species evaluation scope of WildLIFT and, more broadly, to the long tail of understudied species in ecology.
WildLIFT's use of oriented bounding boxes as a species-agnostic geometric abstraction trades articulated detail for taxonomic generality.

\paragraph{Multi-sensor and controlled setups.}
High-fidelity 3D pose recovery has been achieved through synchronised multi-camera arrays~\cite{dunnGeometricDeepLearning2021, 9561338} in laboratory settings.
Field-deployed systems such as WildPose~\cite{muramatsuWildPoseLongrange3D2025} integrate LiDAR with telephoto lenses for long-range 3D tracking, and aerial photogrammetry has enabled body mass and volume estimation of marine vertebrates from drone imagery~\cite{christiansenEstimatingBodyMass2019, hodgson_rapid_2020, stone_using_2025}.
WildLIFT targets a different operational setting: dynamic scenes captured by a single off-the-shelf RGB drone camera, without requiring stationary subjects, calibrated multi-camera rigs, or active sensors such as LiDAR.
This constraint reflects the reality that the majority of archival drone wildlife footage was collected with consumer-grade equipment and cannot be retroactively supplemented with additional sensor data.

\subsection*{Open-vocabulary segmentation: extended context}

The main text describes WildLIFT's use of Grounded-SAM~\cite{ren_grounded_2024} for open-vocabulary instance segmentation.
Here we provide additional context on the constituent models and ecological applications that motivate this choice.

SAM~\cite{Kirillov_2023_ICCV}, trained on 1.1 billion masks, provides class-agnostic segmentation with zero-shot transfer but requires spatial prompts (points or bounding boxes) and lacks semantic understanding of \emph{what} it segments.
Grounding DINO~\cite{liu_grounding_2024} complements SAM by providing open-vocabulary object detection from arbitrary text queries.
Grounded-SAM-2~\cite{ren_grounded_2024} chains these capabilities: Grounding DINO localises objects matching text descriptions, and SAM generates precise masks from the resulting bounding boxes.
SAM~2~\cite{ravi_sam_2024} further extends promptable segmentation to video with memory-based propagation, including explicit demonstrations on wildlife in drone footage.

Domain-adapted vision--language models are also gaining traction in ecology.
BioCLIP~\cite{stevens2024bioclip}, trained on 10 million biological images spanning 450{,}000 taxa, outperforms standard CLIP by 16--17\% on species classification tasks, suggesting that domain-specific foundation models may eventually improve WildLIFT's segmentation front-end.
WildLIFT's modular architecture is designed to accommodate such substitutions as the field advances.

\bibliographystyle{sn-nature}
\bibliography{ref_fin,sn-article-template/ref_fin_nm}